\documentclass[acmtog,nonacm]{acmart}
\acmSubmissionID{1458}

\usepackage{booktabs} 

\usepackage[ruled]{algorithm2e} 

\usepackage{graphicx}
\usepackage{overpic}

\newcommand\blfootnote[1]{%
  \begingroup
  \renewcommand\thefootnote{}%
  \def\@makefntext##1{\noindent##1}
  \footnote{#1}%
  \addtocounter{footnote}{-1}%
  \endgroup
}

\SetAlFnt{\small}
\SetAlCapFnt{\small}
\SetAlCapNameFnt{\small}
\SetAlCapHSkip{0pt}

\acmJournal{TOG}

\begin{document}
\title{Training-free Stylized Text-to-Image Generation with Fast Inference}

\makeatletter
\let\@authorsaddresses\@empty
\makeatother

\author{
Xin Ma\textsuperscript{\rm 1}, 
Yaohui Wang\textsuperscript{\rm 2$\ddagger$}, 
Xinyuan Chen\textsuperscript{\rm 2},
Tien-Tsin Wong\textsuperscript{\rm 1},
Cunjian Chen\textsuperscript{\rm 1$\ddagger$} \\
\small
\textsuperscript{1}Department of Data Science \& AI, Faculty of Information Technology, Monash University
\small
\textsuperscript{2}Shanghai AI Laboratory
}

\renewcommand{\shortauthors}{Xin Ma, Yaohui Wang, Xinyuan Chen, Tien-Tsin Wong and Cunjian Chen}

\begin{abstract}
Although diffusion models exhibit impressive generative capabilities, existing methods for stylized image generation based on these models often require textual inversion or fine-tuning with style images, which is time-consuming and limits the practical applicability of large-scale diffusion models. To address these challenges, we propose a novel stylized image generation method leveraging a pre-trained large-scale diffusion model without requiring fine-tuning or any additional optimization, termed as \textbf{OmniPainter}. Specifically, we exploit the self-consistency property of latent consistency models to extract the representative style statistics from reference style images to guide the stylization process. Additionally, we then introduce the norm mixture of self-attention, which enables the model to query the most relevant style patterns from these statistics for the intermediate output content features. This mechanism also ensures that the stylized results align closely with the distribution of the reference style images. Our qualitative and quantitative experimental results demonstrate that the proposed method outperforms state-of-the-art approaches. The project page is available at~\url{https://maxin-cn.github.io/omnipainter_project}.
\end{abstract}

%
%


%
%


\begin{teaserfigure}
\includegraphics[width=1.0\textwidth]{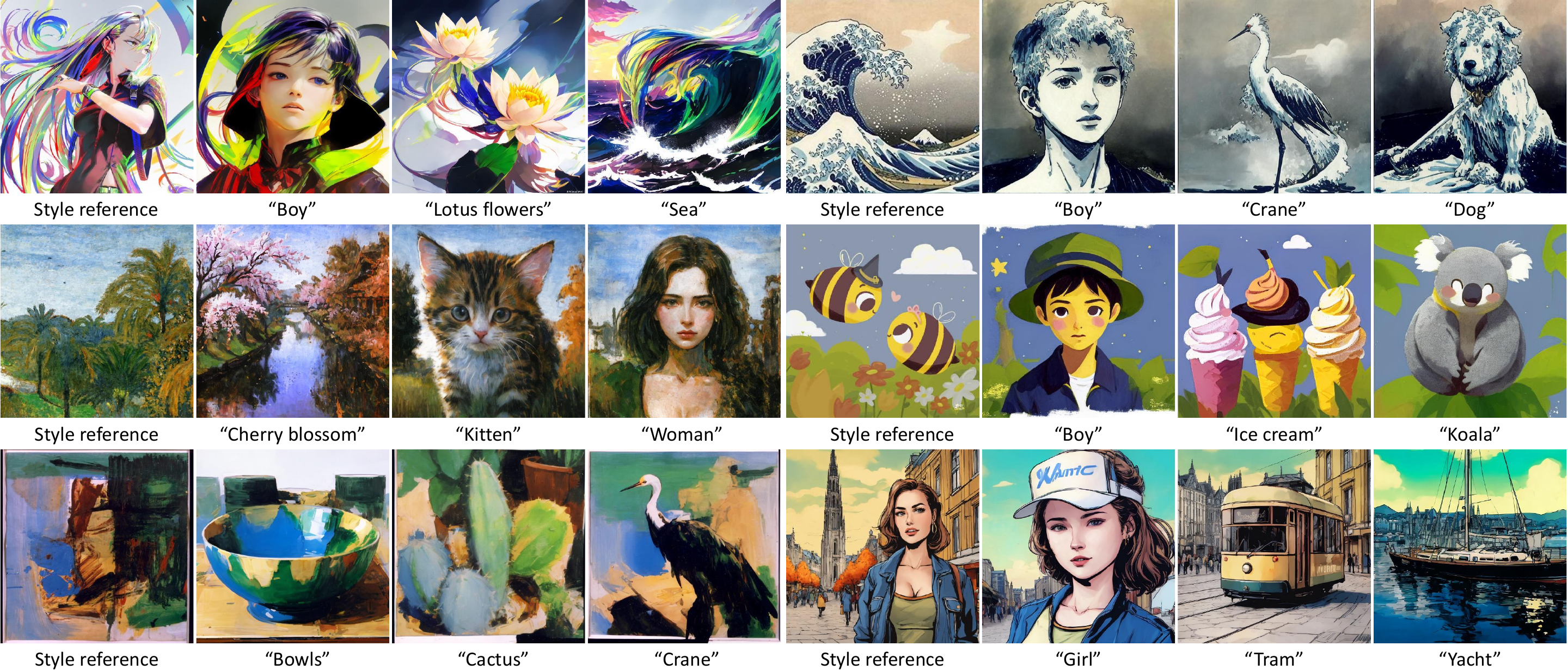}
\caption{\textbf{Examples generated by OmniPainter.} Our method can generate images in desired styles from any textual prompt, requiring only one style image.}
\label{fig:teaser}
\end{teaserfigure}

\maketitle

\blfootnote{$\ddagger$ Corresponding authors.}

\section{Introduction}
\label{sec:intro}

\begin{figure}[!t]
\centering
\begin{tabular}{c@{\hspace{0.3em}}c@{\hspace{0.3em}}c@{\hspace{0.3em}}c@{\hspace{0.3em}}c@{\hspace{0.3em}}c@{\hspace{0.3em}}c}

\includegraphics[width=0.235\linewidth]{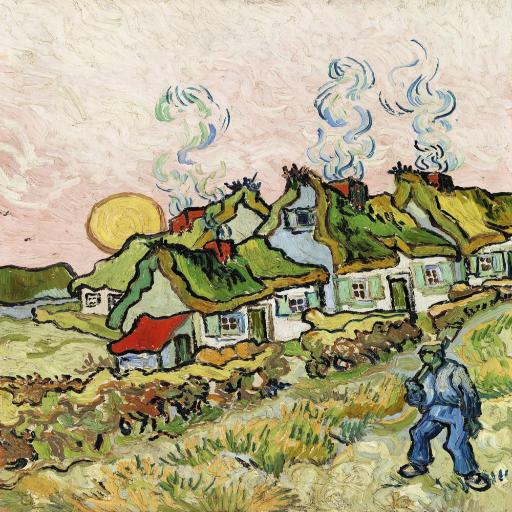} &
\includegraphics[width=0.235\linewidth]{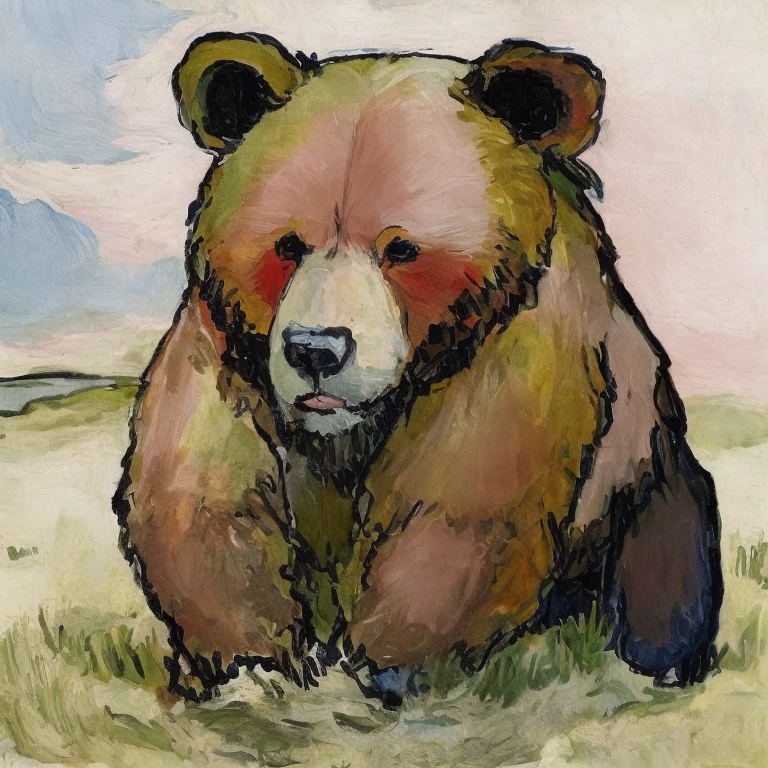} & 
\includegraphics[width=0.235\linewidth]{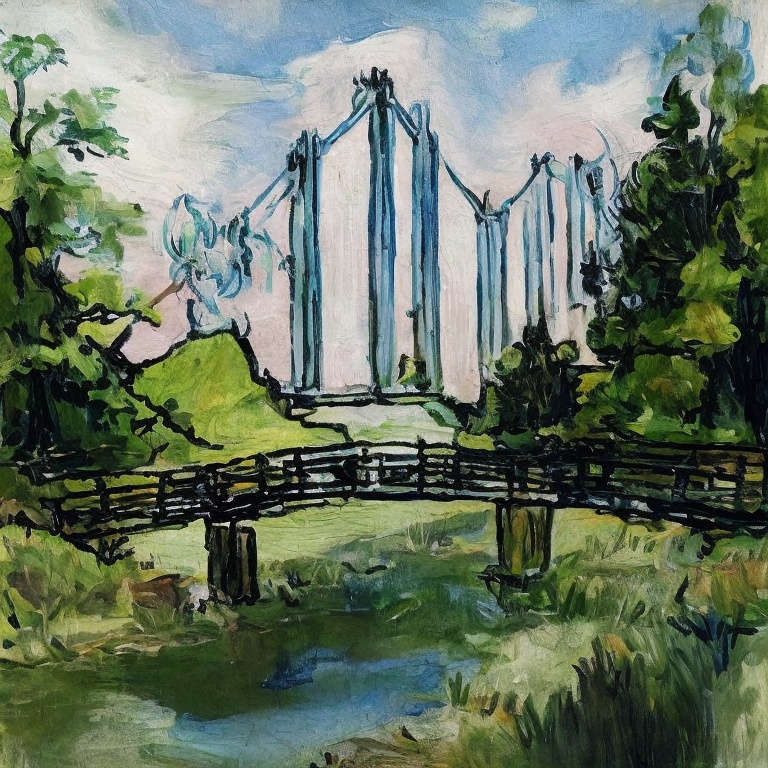} &
\includegraphics[width=0.235\linewidth]{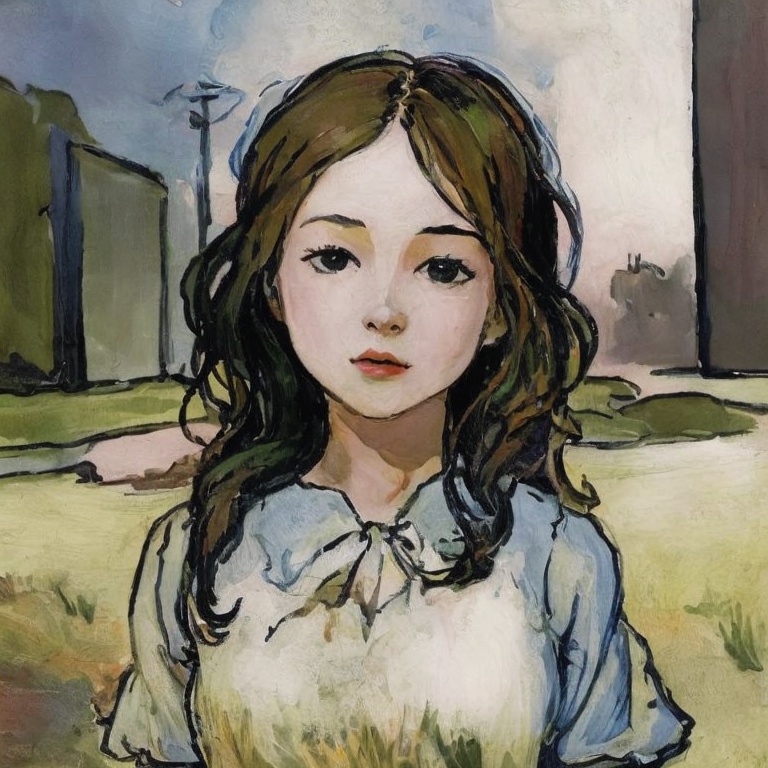}
\\

Style refer & ``Bear'' & ``Bridge'' & ``Girl'' \\



\includegraphics[width=0.235\linewidth]{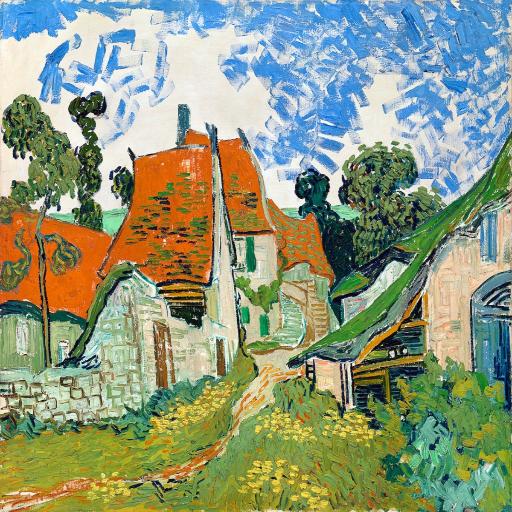} &
\includegraphics[width=0.235\linewidth]{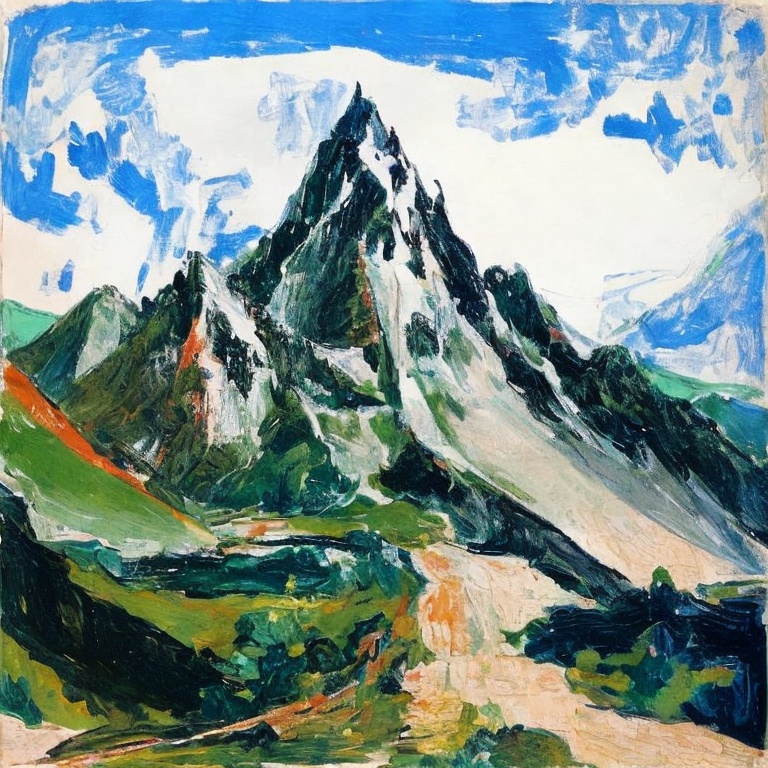} & 
\includegraphics[width=0.235\linewidth]{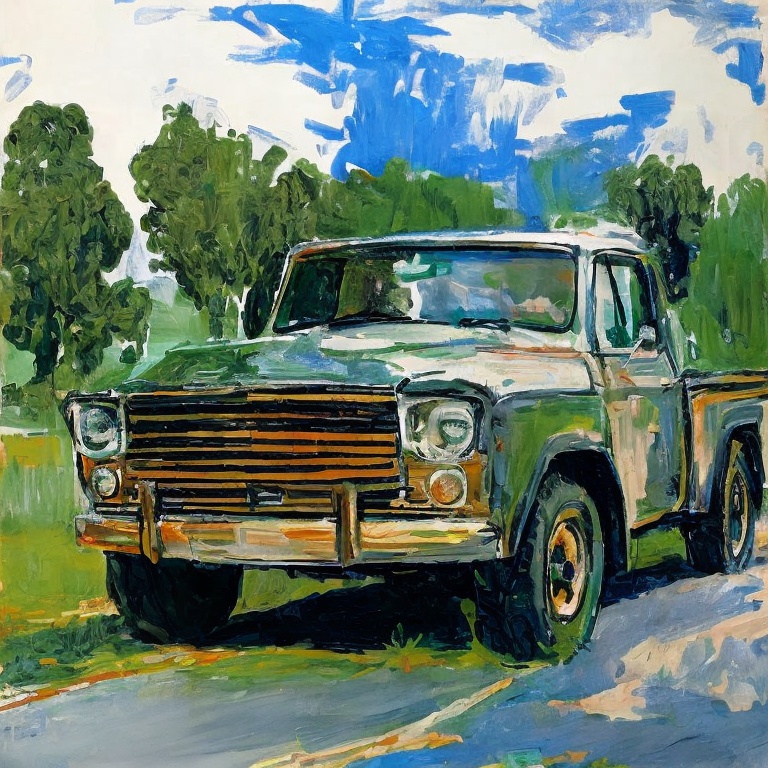} &
\includegraphics[width=0.235\linewidth]{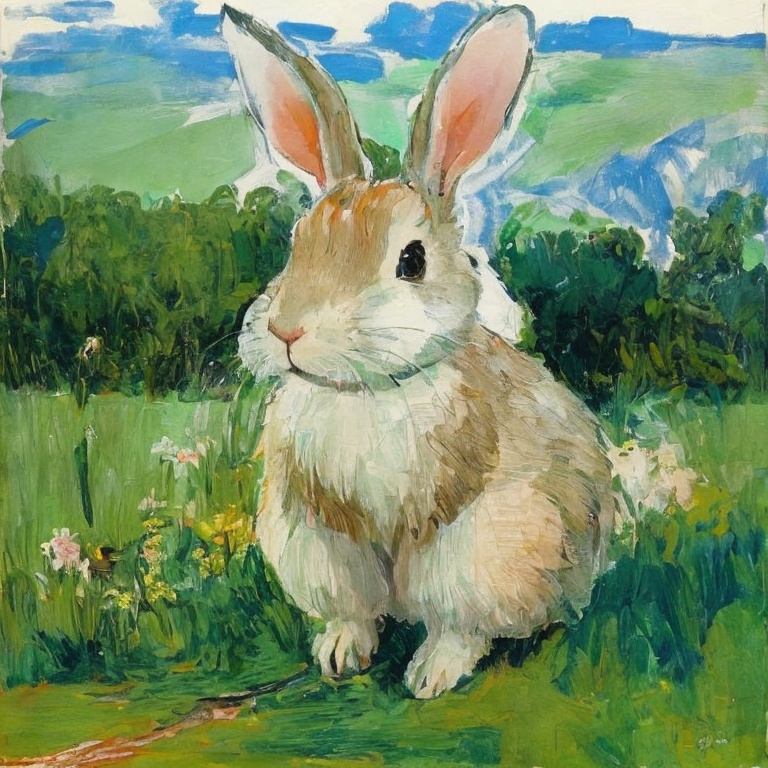}
\\

Style refer & ``Mountain'' & ``Truck'' & ``Rabbit'' \\

\end{tabular}
\vspace{-0.3cm}
\caption{Examples generated by our method using different paintings of Van Gogh as style images.}
\label{fig_van_gogh}
\vspace{-0.3cm}
\end{figure}

Text-to-image (T2I) diffusion models trained on large-scale datasets have demonstrated remarkable capability in generating diverse, detail-rich, high-quality images across a wide range of genres and themes~\cite{chang2023muse,rombach2022high,saharia2022photorealistic,chen-pixart,chen2024pixart}. 
Existing models can produce images with styles specified by users through text prompts (e.g. ``oil painting,'' or ``watercolor''), as the training data contains examples of major styles. However, if a style is less popular or more finely categorized, generating the desired style with nuanced brushwork and unique color schemes becomes challenging, even with extensive prompt engineering efforts~\cite{gal2022stylegan,cui2025instastyle}. 
For instance, an artist like Van Gogh can have multiple styles throughout his career (Fig.~\ref{fig_van_gogh}), yet his paintings are likely labeled simply as “Van Gogh” in the training data. In other words, providing a reference image of the desired style should be a more effective means of describing and controlling the nuanced brushwork, color scheme, and other subtle characteristics to guide the generation process. 

\begin{figure}[!t]
\centering
\begin{tabular}{c@{\hspace{0.3em}}c@{\hspace{0.3em}}c@{\hspace{0.3em}}c@{\hspace{0.3em}}c@{\hspace{0.3em}}c@{\hspace{0.3em}}c}

\includegraphics[width=0.235\linewidth]{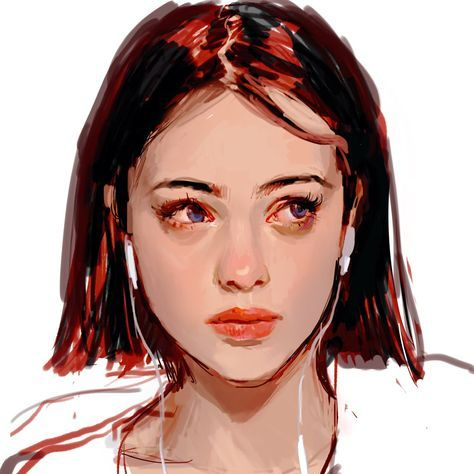} &
\includegraphics[width=0.235\linewidth]{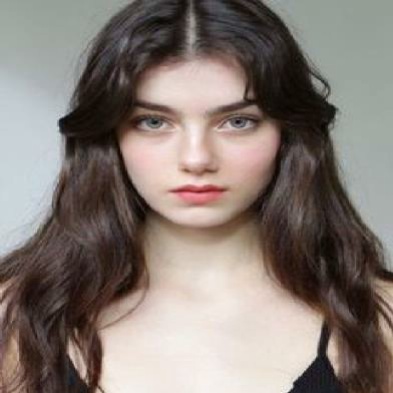} & 
\includegraphics[width=0.235\linewidth]{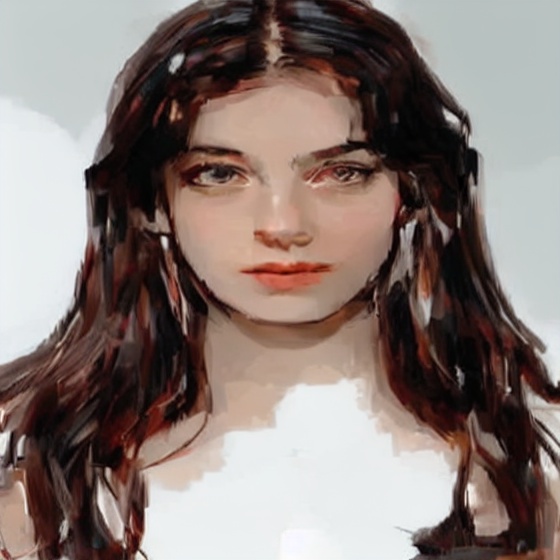} &
\includegraphics[width=0.235\linewidth]{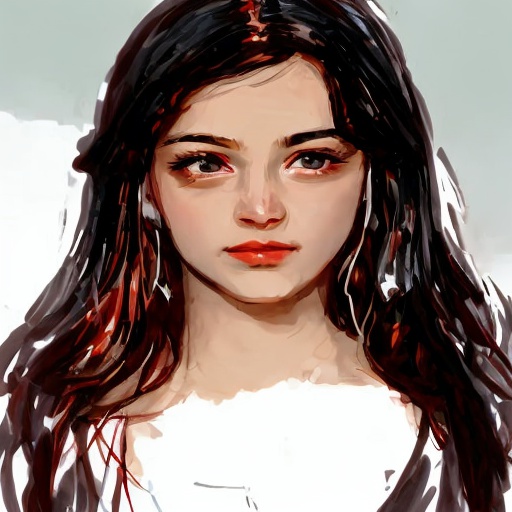}
\\

\includegraphics[width=0.235\linewidth]{images/naive_method_issue/woman_style.jpg} &
\includegraphics[width=0.235\linewidth]{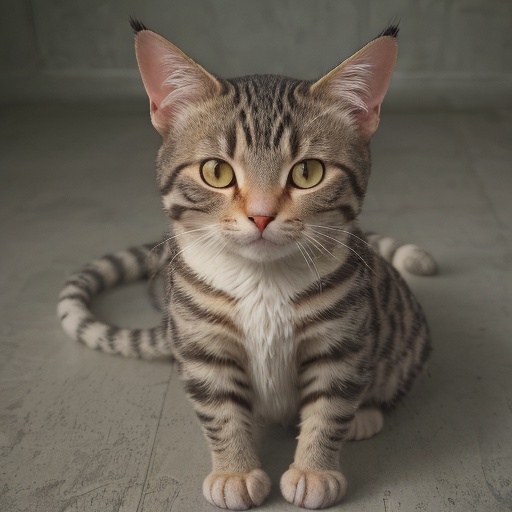} & 
\includegraphics[width=0.235\linewidth]{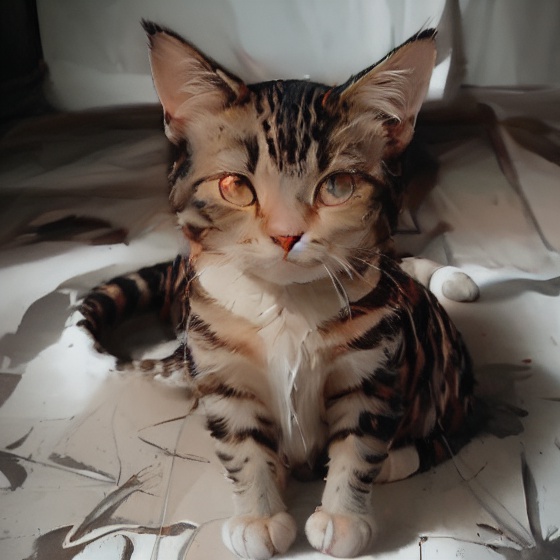} &
\includegraphics[width=0.235\linewidth]{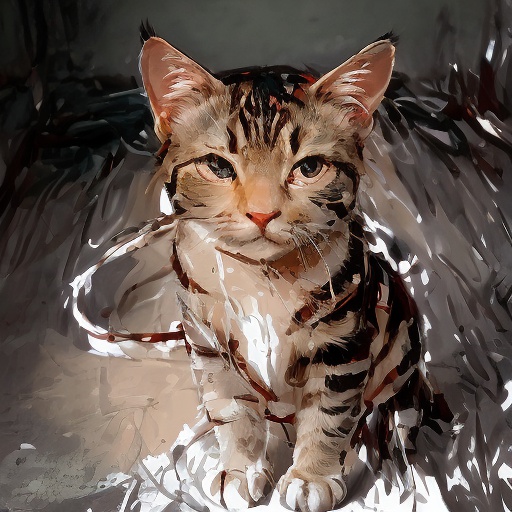}
\\

\includegraphics[width=0.235\linewidth]{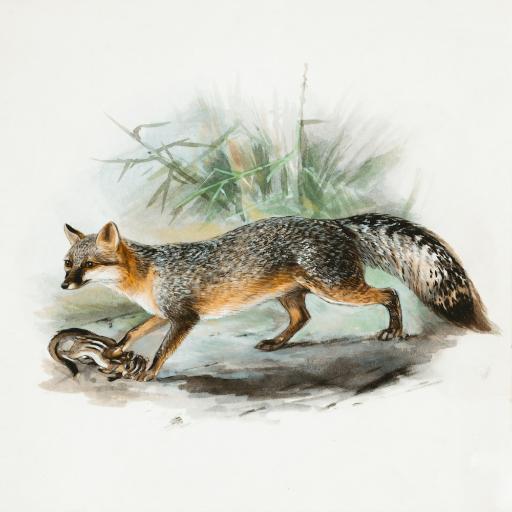} &
\includegraphics[width=0.235\linewidth]{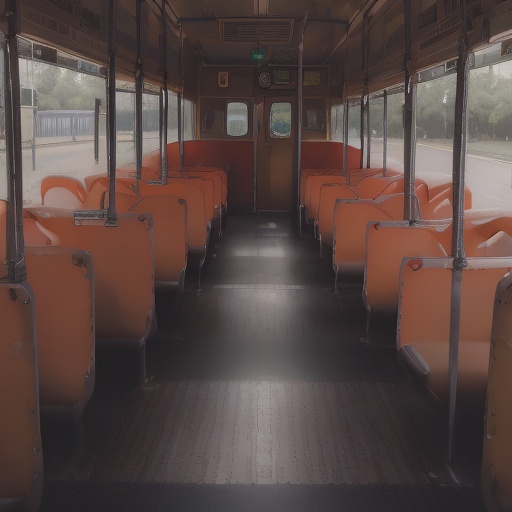} & 
\includegraphics[width=0.235\linewidth]{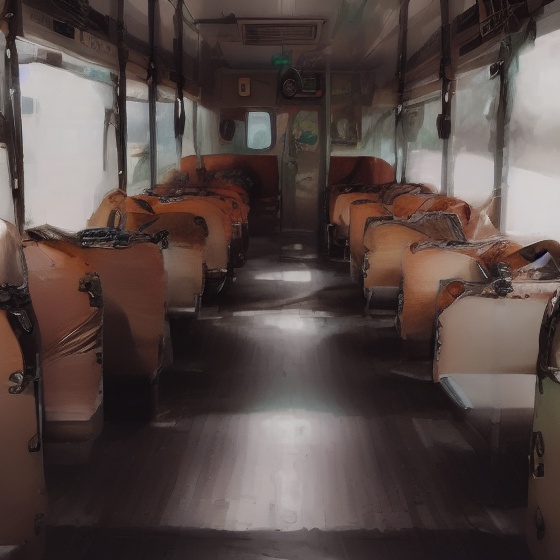} &
\includegraphics[width=0.235\linewidth]{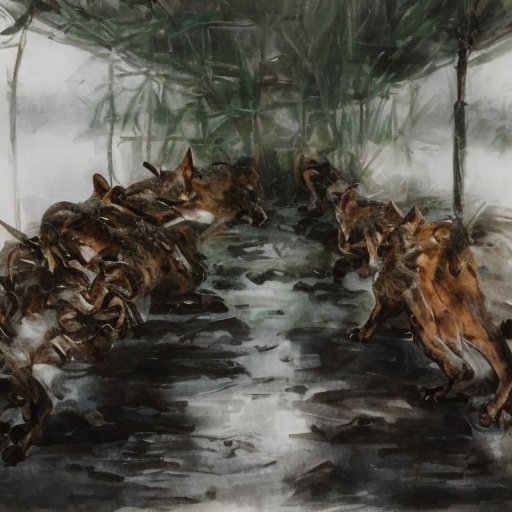}
\\

Style refer & Content & Z-STAR & ZePo \\

\end{tabular}
\vspace{-0.3cm}
\caption{\textbf{Examples of using the style transfer method for stylized T2I generation directly.} We first generate images from prompts using the T2I method~\cite{luo2023latent}, then apply style transfer methods~\cite{deng2023z,liu2024zepo} to incorporate the specified style.}
\label{fig_naive_method_issue}
\vspace{-0.3cm}
\end{figure}

A straightforward approach is to first generate the image with a pre-trained T2I model given the content prompt. Then the generated image is converted to a specific style using the state-of-the-art style transfer method given a reference style image. Fig.~\ref{fig_naive_method_issue} (top row) shows one such success example. However, it mainly works when the reference style image and content image share certain similarities. This approach may fail when the reference style image and content image are substantially different like the examples in Fig.~\ref{fig_naive_method_issue} (middle \& bottom rows). An alternative approach is to take a few more reference style images and minimally train/fine-tune the models using LoRA or adaptor~\cite{hu2021lora,houlsby2019parameter,sohn2023styledrop,cui2025instastyle}, or even fine-tune the entire T2I models~\cite{everaert2023diffusion}. 
However, these methods require more time and effort, making them less convenient than simply providing a reference style image.

In this paper, we present \textbf{OmniPainter}, a fast, training-free, and inversion-free stylized T2I generation method that addresses the aforementioned challenges. Our core idea is to retain the high-level semantic content from the original T2I generation process while incorporating representative style statistics (typically lower-level features) from the reference style image. 
To achieve this, we extract representative style statistics from the reference style image and incorporate them through a pseudo cross-attention mechanism during the denoising process. While a straightforward approach to obtain these statistics could involve DDIM Inversion~\cite{wallace2023edict}, which generates signature keys and values to capture style, this method incurs significant computational overhead due to the additional inversion step.
To enable fast stylized image synthesis, we propose to build our method on top of latent consistency models (LCMs)~\cite{song2023consistency,luo2023latent}, instead of the original diffusion models. Leveraging the inherent self-consistency property of LCMs, we can extract representative style statistics (in the form of key and value statistics) directly from the reference style image without requiring DDIM Inversion. Then, during the image generation, implemented as LCM but with a few-step denoising, we apply a mixture-of-self-attention (MSA) mechanism to guide the generation process with these style statistics. Our method operates on the content features within self-attention spaces so as to retain the original content guided by the text prompt.
To further mitigate the distribution discrepancy between the generated images and the style images, particularly in terms of color deviations, we introduce an AdaIN~\cite{saharia2022photorealistic} operation before MSA, which aims to align their distributions more effectively. We term this the norm-mixture-of-self-attention (NMSA). With the proposed method, as shown in Fig.~\ref{fig:teaser}, we can seamlessly integrate the representative style statistics during the T2I generation process, so that the generated image aligns well to both the given prompt and the style reference.

Our OmniPainter generates stylized images in six sampling steps, significantly reducing the computation time and enhancing the practicality of diffusion-based text-to-image stylization. 
Extensive experiments have been conducted to validate the effectiveness of our method. As shown in Fig.~\ref{fig_performance_vs_efficiency}, our method takes only 0.7 seconds on average to generate images, and achieves the highest style score and comparable content fidelity scores. To summarize, our contributions are as follows:
\begin{itemize}
\item We introduce OmniPainter, a new fast, training-free, and inversion-free stylized T2I generation method that requires only 0.7 seconds on average to synthesize a high-quality image with the desired style.
\item We introduce the norm mixture of self-attention mechanism to seamlessly integrate the representative style statistics into the generation of images that remain aligned well with the text prompt. 
\end{itemize}
\section{Related Works}

\begin{figure*}[!ht]
\centering
\includegraphics[width=1.0\linewidth]{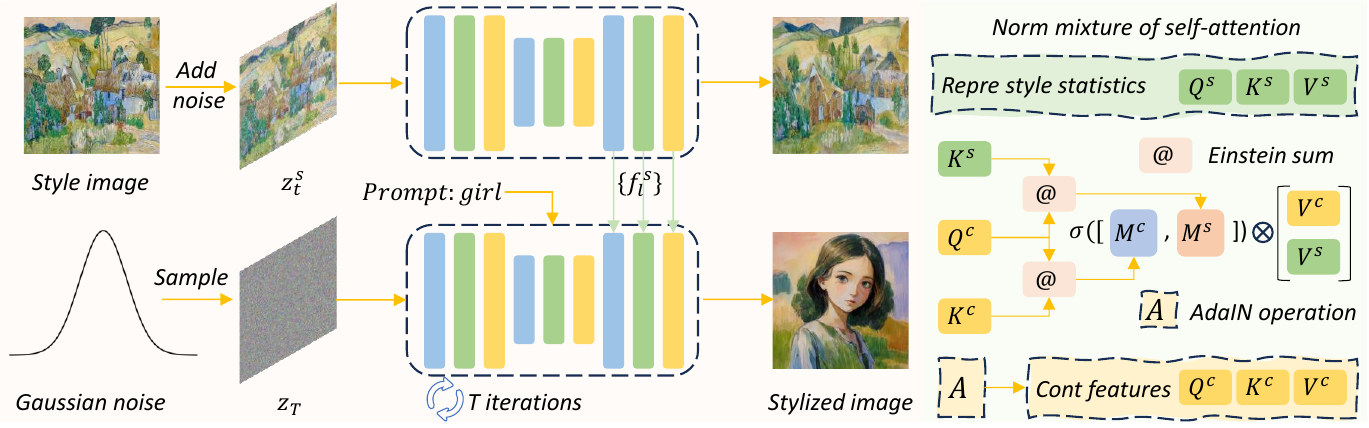}
\vspace{-0.3cm}
\caption{\textbf{The overall pipeline of our method.} Here, $\sigma$, ``Repre style statistics'', and ``Cont features'' are the softmax operation, representative style statistics, and content features, respectively. The whole stylization process operates in the latent space of the pre-trained VAE.}
\label{fig_pipeline}
\vspace{-0.3cm}
\end{figure*}

\textbf{Personalized text-to-image Synthesis.} Text-to-image generation has recently become a widely discussed topic~\cite{rombach2022high,podell2023sdxl,esser2024scaling,chen-pixart,chen2024pixart,ma2022contrastive,ma2021free,ma2024uncertainty,luo2022style}, due to its remarkable generalization capabilities demonstrated by pre-trained vision-language models~\cite{radford2021learning,wang2023learning,ma2024gerea,ma2024drvideo,li2025hier,li2024hi}, diffusion models~\cite{song2020denoising,ho2020denoising,song2021score,wang2024lavie,ma2024latte,ma2024cinemo,chen2023seine}, and auto-regressive models~\cite{tian2024visual,chang2023muse}. 
Several personalized text-to-image synthesis methods aiming at incorporating personal assets have been proposed, leveraging the powerful pre-trained text-to-image models. Textual inversion~\cite{gal2022image} and Hard Prompts Made Easy~\cite{wen2024hard} identify text representations (e.g., embeddings, tokens) corresponding to a set of images of a specific object, enabling personalized T2I generation without modifying the parameters of the pre-trained T2I model. Prompt-to-prompt~\cite{hertz2022prompt} leverages these characteristics by replacing or re-weighting the attention maps between text prompts and their corresponding edited images. Furthermore, Null-text Inversion~\cite{mokady2023null} identifies that the classifier-free guidance in conditional text-to-image generation amplifies the cumulative errors at each DDIM inversion step~\cite{wallace2023edict}. To address this, it introduces null-text optimization, building on the Prompt-to-Prompt framework, which enables real image editing capabilities. 
Plug-and-Play~\cite{tumanyan2023plug} and MasaCtrl~\cite{cao2023masactrl} shift the emphasis from text prompts to spatial features, utilizing the self-attention mechanisms within the U-Net architecture of the latent diffusion model to inject the characteristics of the input image into the target generation branch.

Other methods involve fine-tuning either partial or all parameters of pre-trained T2I models~\cite{wang2024instantstyle}. DreamBooth~\cite{ruiz2023dreambooth} fine-tunes the whole text-to-image model using just a few images of the subject of interest. This enables it to be more expressive and capture the subject with enhanced detail and fidelity. To save computational resources, more parameter-efficient fine-tuning methods, such as LoRA~\cite{hu2021lora,guo2023i2v} or adapter tuning~\cite{houlsby2019parameter}, are introduced. ZipLoRA~\cite{shah2025ziplora} introduces an efficient approach to merge independently trained style and subject LoRAs, enabling the generation of any user-defined subject in any desired style. Similarly, T2I-Adapter~\cite{mou2024t2i} learns simple and lightweight adapters to align the internal knowledge in pre-trained T2I models with the external control signals while freezing the original large pre-trained T2I models. Our proposed stylized T2I generation method falls into the category of personalized T2I synthesis, but it does {\em not} require any fine-tuning or the DDIM inversion.

\noindent\textbf{Neural Style Transfer} seeks to create an image that preserves the content structure of the content image while adopting the artistic style of the style image, leveraging the power of deep neural networks~\cite{gatys2016image,huang2017arbitrary,chen2022quality}.
Gatys \textit{et al.}~\cite{gatys2016image} find that Gram matrices of image features extracted from pre-trained VGG models can effectively represent style. They proposed an optimization-based approach to generate stylized images by minimizing the differences between the Gram matrices of the generated and style images. AdaIN~\cite{huang2017arbitrary} employs an adaptive instance normalization technique to align the mean and variance of the content image with those of the style image, enabling global style transfer. Li \textit{et al.}~\cite{li2017universal} proposes utilizing feature transformations, specifically whitening and coloring, to directly align the statistical properties of content features with those of a style image within the deep feature space. Some methods, such as SANet~\cite{deng2020arbitrary}, MAST~\cite{park2019arbitrary}, and AdaAttn~\cite{liu2021adaattn}, leverage attention mechanisms between content and style features to infuse appropriate stylistic patterns into the content effectively. Other approaches exploit the ability of Transformers to capture long-range features, further enhancing the quality of stylized results~\cite{deng2022stytr2,wu2021styleformer,wei2022comparative,tang2023master,wang2022fine,zhang2024s2wat,vaswani2017attention,liu2024dual}.

Recently, T2I models trained on large paired text-to-image datasets have demonstrated remarkable zero-shot generation capabilities. As a result, many efforts have leveraged this advantage for neural style transfer. Style Injection in Diffusion~\cite{chung2024style}, Z-STAR~\cite{deng2023z}, and Z-STAR+~\cite{deng2024z} adopt a similar approach. They first project the content image and style image into Gaussian noise using DDIM inversion. Then, they inject the style characteristics into the key and value of the content features within the self-attention mechanism of Stable Diffusion. Our proposed stylized T2I generation method shares the spirit of style injection, but the injection is incorporated during the T2I generation, no content images are needed in the first place. 

\noindent\textbf{Stylized Image Generation} aims to create images in a desired style based on a few style images, representing a new paradigm in image generation. Although stylized image generation is similar to the neural style transfer task mentioned above, they are fundamentally different. Neural style transfer takes two input images (the content and the style images) and addresses an image translation task, focusing on transferring the artistic style of the style image onto the content image. On the contrary, stylized image generation creates images with a specific style conditioned on the given prompts. Diffusion in Style~\cite{everaert2023diffusion} adjusts the initial latent distribution using the mean and variance of a set of style images, followed by fine-tuning Stable Diffusion (SD)~\cite{rombach2022high} on this style image set to improve the stylized results. Similarly, InstaStyle~\cite{cui2025instastyle} first uses the inverted initial noise of the style image to generate a set of images with a similar style pattern, then fine-tunes SD using LoRA and the learned style token. 
StyleDrop~\cite{sohn2023styledrop} adopts a training procedure similar to InstaStyle, training an adapter for each individual style pattern using paired text and images. In this approach, the textual prompts describe both the content and style of the given images. Unlike the above methods requiring LoRA, adaptor, or finetuning, our proposed method does not require any training/finetuning or any parameter-efficient tuning but achieves real-time stylized image generation aligning to both the textual prompts and the reference style images. 

\section{Methodology}


\subsection{Preliminaries}
\textbf{Latent Diffusion Models} (LDMs) are efficient diffusion models that employ the diffusion process in the low-dimensional latent space of the pre-trained VAE rather than the high-dimensional pixel space~\cite{song2020denoising,rombach2022high,ho2020denoising,kingma2013auto,kingma2019introduction}. An encoder $\mathcal{E}$ of the pre-trained VAE is firstly utilized in LDMs to project the input data sample $x \in p_{\rm data}$ into a low-dimensional latent code $z = \mathcal{E}(x)$. The data distribution is then learned according to two key processes: diffusion and denoising. The diffusion process generates the perturbed sample $z_{t}$ by the following formulation,
\begin{equation}
z_t = {\sqrt{\overline{\alpha}_{t}}}z + \sqrt{1-{\overline{\alpha}_{t}}}\epsilon,
\label{equ_diffusion_forward}
\end{equation}
where $\epsilon\sim \mathcal{N}(0,1)$ and this process gradually adds Gaussian noise to the latent code $z$. Note that $\overline{\alpha}_{t}$ and $t$ are the pre-defined noise scheduler and the diffusion timestep, respectively. The denoising process is the inversion of the diffusion process, which learns to predict a less noisy sample $z_{t-1}$: $p_\theta(z_{t-1}|z_t)=\mathcal{N}(\mu_\theta(z_t),{\Sigma_\theta}(z_t))$ and make the variational lower bound of log-likelihood reduce to $\mathcal{L_\theta}=-\log{p(z_0|z_1)}+\sum_tD_{KL}((q(z_{t-1}|z_t,z_0)||p_\theta(z_{t-1}|z_t))$. In this context, $\mu_\theta$ means a denoising model and is trained with the following objective,
\begin{equation}
\label{equ_l_simple}
\mathcal{L}_{\rm simple} = \mathbb{E}_{\mathbf{z}\sim p(z),\ \epsilon \sim \mathcal{N} (0,1),\ t}\left [ \left \| \epsilon - \epsilon_{\theta}(\mathbf{z}_t, t)\right \|^{2}_{2}\right].
\end{equation}

\noindent\textbf{Latent Consistency Models} (LCMs) are based on a concept similar to that of LDMs, operating within the low-dimensional space of a pre-trained VAE~\cite{song2023consistency, song2023improved, luo2023latent}. LCMs have demonstrated significant potential as a new class of generative models, offering faster sampling while maintaining high generation quality. In LCMs, the consistency function $f_\theta(z_t, t)$ guarantees that each anchor point $z_t$ along the sampling trajectory can be precisely mapped back to the initial latent code $z_0$, thus maintaining self-consistency within the model. The consistency function is mathematically defined as the following formulation,
\begin{equation}
    f_\theta(x, t) = c_{\rm skip}(t)x + c_{\rm out}(t)F_\theta(x,t),
\end{equation}
where $c_{\rm skip}$ and $c_{\rm out}$ are the differentiable functions to ensure the differentiability of $f_\theta(x, t)$, subject to the conditions $c_{\rm skip}(0)=1$ and $c_{\rm out}(0)=0$. $F_\theta(x,t)$ is a deep neural network model. The self-consistency characteristic of $f(x, t)$ is ensured by the following optimization objective,
\begin{equation}
    \min_{\theta, \theta^-; \phi} \mathbb{E}_{z_0, t} \left[ d\left(f_{\theta}(z_{t+1}, t+1), f_{\theta^-}(\hat{z}_{t}^\phi, t)\right) \right],
\label{equ_lcm_objective}
\end{equation}
where ${\theta^-}$ is updated through the exponential moving average (EMA) of the parameter $\theta$ we intend to learn, i.e., $\theta^- \leftarrow \mu\theta^- + (1 - \mu)\theta$. And $d(\cdot,\cdot)$ is a metric function that measures the distance between two samples, e.g., the squared $\mathcal{L}_2$ distance $d(x, y)=||x - y||^2$. $\hat{z}_{t}^\phi$ is a one-step estimation of $z_t$ from $z_{t+1}$ by using the following formulation,
\begin{equation}
    \hat{z}_{t}^\phi = z_{t+1} + \Delta t \Phi(z_{t+1}, t+1;\phi),
\end{equation}
where $\Phi$ represents the one-step ordinary differential equation (ODE) solver~\cite{song2023consistency,karras2022elucidating}. 


\subsection{Pipeline Overview}

In this paper, we extend LCMs to enable them with the stylized T2I generation capability. Our pipeline (Fig.~\ref{fig_pipeline}) generates a stylized image from a style image $I^s$ and a prompt $p$. The upper branch extracts representative style statistics from $I^s$ using one-step denoising with LCMs (Sec.~\ref{sec:extracting_representative_style_statistics}). The lower branch uses these statistics and $p$ in a few-step T2I denoising process to produce the stylized image. The statistics are seamlessly integrated via NMSA (Sec.~\ref{sec_mixture_of_style_distribution_normalized_self-attention}).


\begin{figure}
\centering
\includegraphics[width=1.0\linewidth]{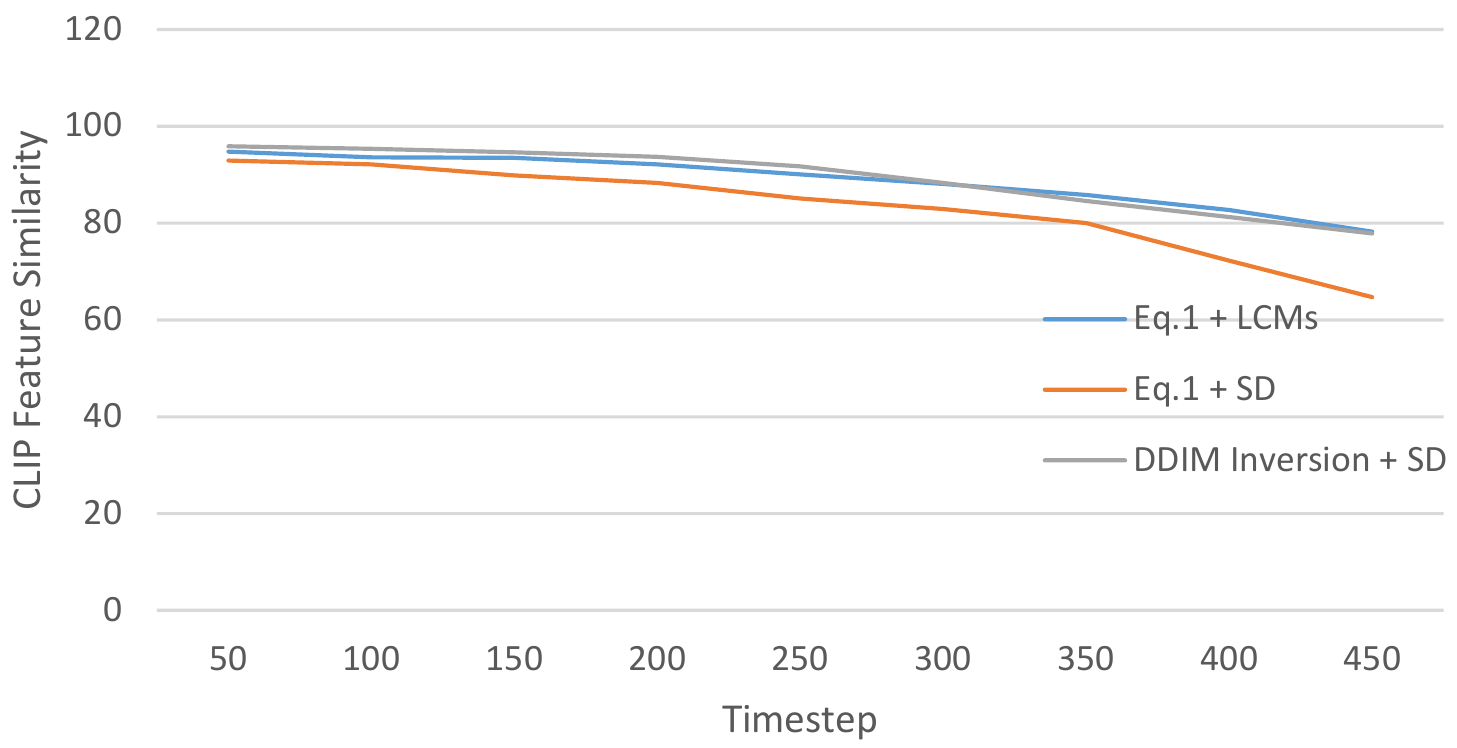}
\vspace{-0.3cm}
\caption{CLIP features similarity of different combinations at different timesteps.}
\label{fig_one_timestep_denosing_with_different_methods}
\vspace{-0.3cm}
\end{figure}

\subsection{Extracting Representative Style Statistics}
\label{sec:extracting_representative_style_statistics}
In Fig.~\ref{fig_one_timestep_denosing_with_different_methods}, we apply two noise injection methods (Eq.~\ref{equ_diffusion_forward} and DDIM Inversion) to the style images and perform a single de-noising step using two different models (LCMs and SD). We extract features from both the clean style images and the denoised images using the image encoder of CLIP and compute their cosine similarity~\cite{radford2021learning,cui2025instastyle}. This figure shows that as the timestep increases, the performance of all three combinations gradually declines. However, the performance gap between the Eq.~\ref{equ_diffusion_forward} + LCMs combination and the DDIM Inversion + SD combination remains relatively small. This indicates that LCMs can effectively extract representative style statistics from noisy style images, which can avoid time-consuming DDIM Inversion operations. The underlying reason for this phenomenon is the optimization objective of LCMs (Eq.~\ref{equ_lcm_objective}) minimizes the difference between the outputs of the consistency function in neighboring samples. This mechanism inherently preserves the representative style statistics even during single step predictions. The denoised style images can be seen in Fig.~\ref{fig_vis_one_timestep_denosing}.




Given a reference style image $I^s$, the encoder $\mathcal{E}$ of the pre-trained VAE is initially used to encode $I^s$ into the latent code $z^s$. Subsequently, the Gaussian noise is added to $z^s$ by using Eq.~\ref{equ_diffusion_forward},
\begin{equation}
 z_t^s = {\sqrt{\overline{\alpha}_{t}}}z^s + \sqrt{1-{\overline{\alpha}_{t}}}\epsilon.
\end{equation}
Finally, the noisy latent code $z_t^s$ is fed into the noise prediction backbone of the LCMs, where the representative style statistics are extracted at each Transformer layer $l$ within the backbone,
\begin{equation}
\{f_l^s\} = F_\theta(z_t^s, t, p).
\end{equation}

\subsection{Norm Mixture of Self-Attention}
\label{sec_mixture_of_style_distribution_normalized_self-attention}
Similar to the definition above, the features derived at each Transformer layer $l$ within the backbone conditional on the given prompt are defined as content features $f_l^c$, which can be written as follows,
\begin{equation}
 \{f_l^c\} = F_\theta(z_t^c, t, p).
\end{equation}

Our objective is to seamlessly integrate the representative style statistics $\{f_l^s\}$ into content features $\{f_l^c\}$ at layer $l$, leading to stylized content feature $\{\hat{f}^c_l\}$. After passing through the Transformer layer in the backbone, the features $[f_l^s, f_l^c]$ are individually mapped to the query $[Q^s_l, Q^c_l]$, key $[K^s_l, K^c_l]$, and value $[V^s_l, V^c_l]$ features within the self-attention module. This manipulation is inspired by the process of Z-STAR~\cite{deng2023z} and ZePo~\cite{liu2024zepo} in style transfer tasks. From now on, we focus on discussing the proposed norm mixture of self-attention to achieve this integration.

\begin{figure}[!tb]
\centering
\begin{tabular}{c@{\hspace{0.3em}}c@{\hspace{0.3em}}c@{\hspace{0.3em}}c@{\hspace{0.3em}}c@{\hspace{0.3em}}c@{\hspace{0.3em}}c}

\includegraphics[width=0.30\linewidth]{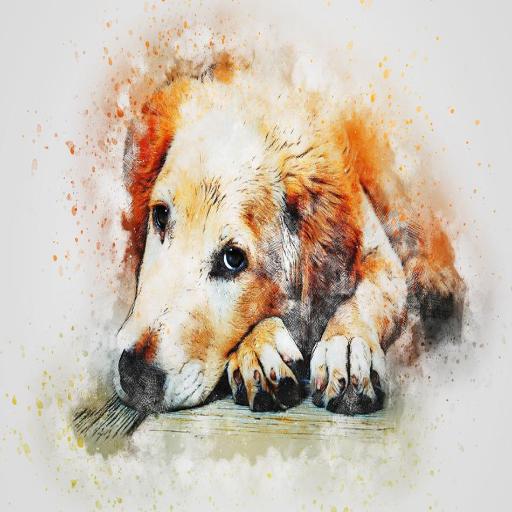} &
\includegraphics[width=0.30\linewidth]{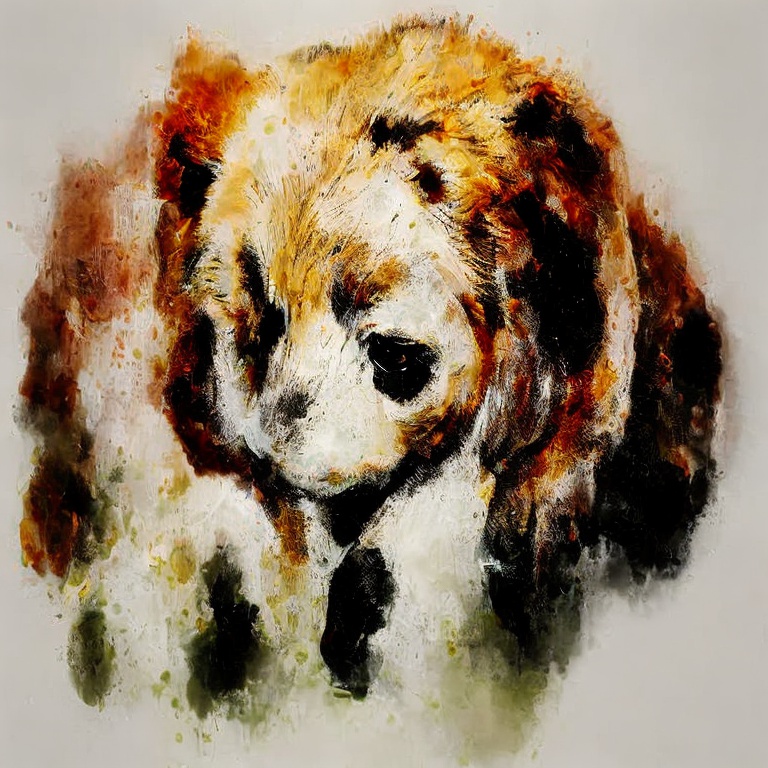} & 
\includegraphics[width=0.30\linewidth]{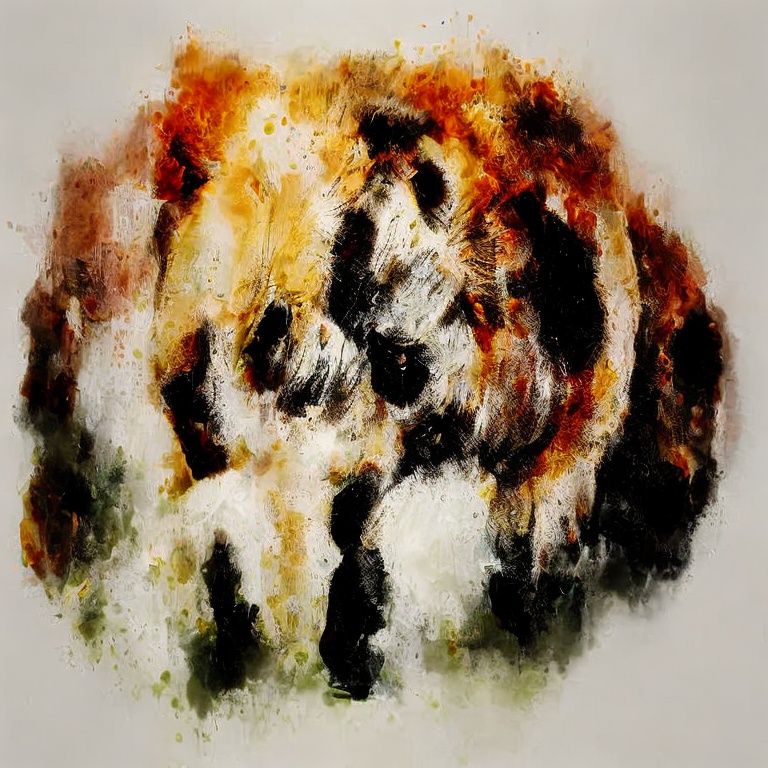}
 
 \\
 
Style refer & ``Girl'' & ``House'' 

\\

\includegraphics[width=0.30\linewidth]{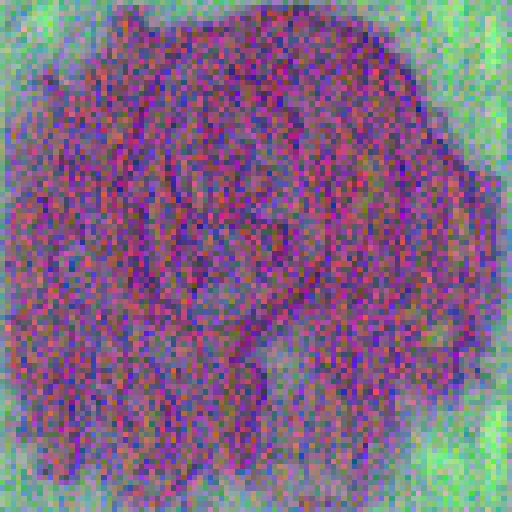} &
\includegraphics[width=0.30\linewidth]{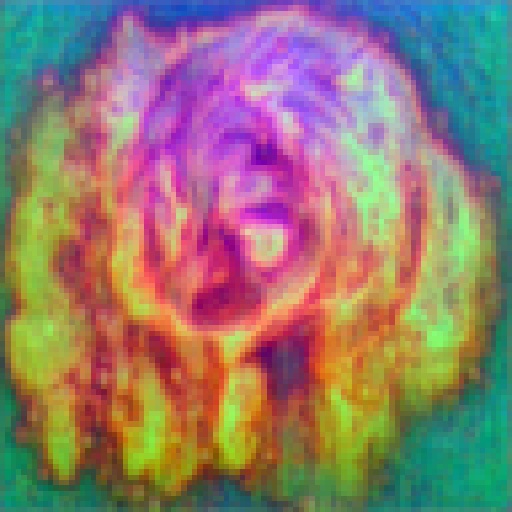} & 
\includegraphics[width=0.30\linewidth]{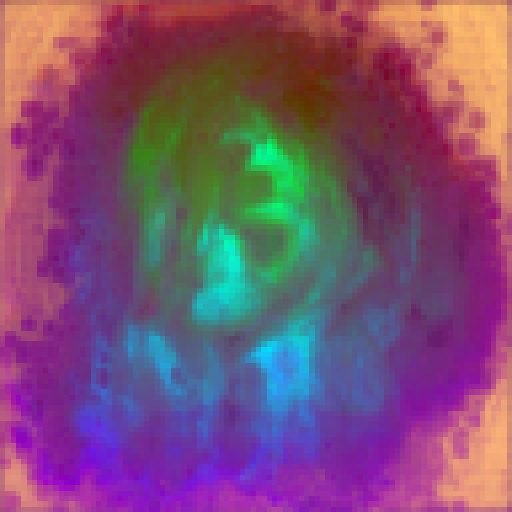}

\\

Resnet block & Query & Self-attn map

\end{tabular}
\vspace{-0.3cm}
\caption{Issues of the direct replacing method and visualization of the top three leading components.}
\vspace{-0.3cm}
\label{fig_method_1}
\end{figure}

\noindent\textbf{Direct Replacement.} It is intuitive that the query can be used to represent semantic information, such as image layout, while the key and value features are used to represent style statistics, such as color, texture, and illumination. Thus, the content features can be used to query the style statistics from the reference style images that best match the input patch. Formally, stylized content feature $\hat{f}^c_l$ can be obtained as follows,
\begin{equation}
\hat{f}^c_l = {\rm \mathcal{A}}(Q^c_l, K^s_l, V^s_l) = \sigma(\frac{Q^c_l{K^s_l}^T}{\sqrt{d}})V^s_l,
\label{eq_method_1}
\end{equation}
where $\mathcal{A}$ and $\sigma$ represent the attention operation and softmax activation function, respectively.
While the approach is straightforward, we observe that it tends to prioritize style statistics at the expense of the semantic information derived from the prompt. For example, as shown in the first row of Fig.~\ref{fig_method_1}, it fails to generate the desired contents corresponding to prompts ``girl" and ``house". We perform PCA on the features or attention maps from various backbone layers~\cite{tumanyan2023plug}, including the ResNet block and the query and key layers of self-attention. The first three principal components are visualized in the second row of Fig.~\ref{fig_method_1}, which further reveals that the semantic information is significantly closer to that of the style reference rather than to the given conditional prompt.

\begin{figure}[!tb]
\centering
\begin{tabular}{c@{\hspace{0.3em}}c@{\hspace{0.3em}}c@{\hspace{0.3em}}c@{\hspace{0.3em}}c@{\hspace{0.3em}}c@{\hspace{0.3em}}c}

\includegraphics[width=0.30\linewidth]{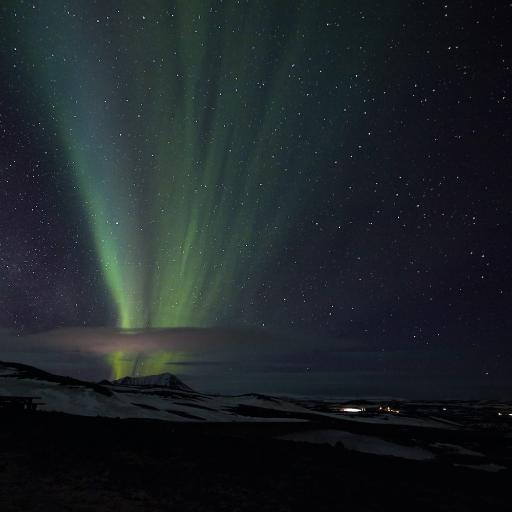} &
\includegraphics[width=0.30\linewidth]{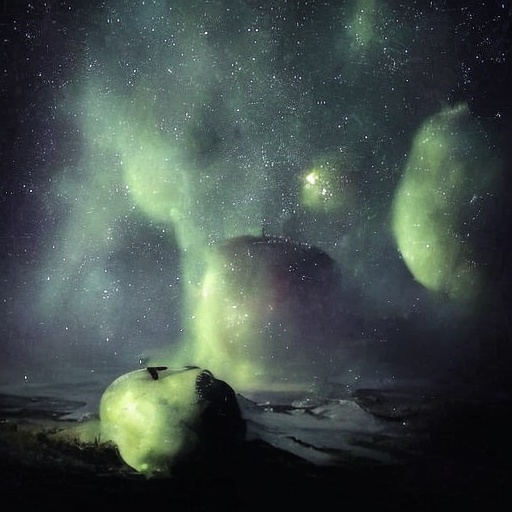} & 
\includegraphics[width=0.30\linewidth]{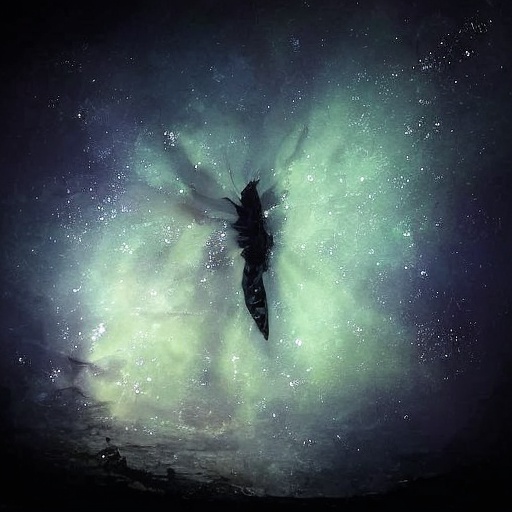}
 
 \\
 
\includegraphics[width=0.30\linewidth]{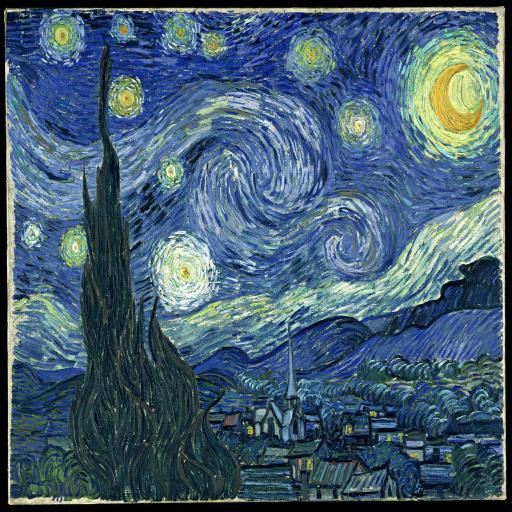} &
\includegraphics[width=0.30\linewidth]{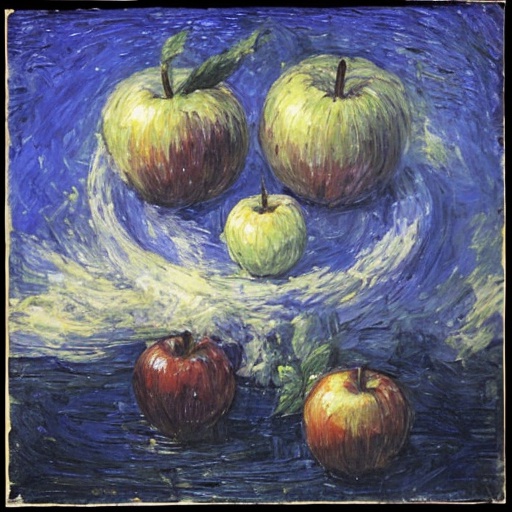} & 
\includegraphics[width=0.30\linewidth]{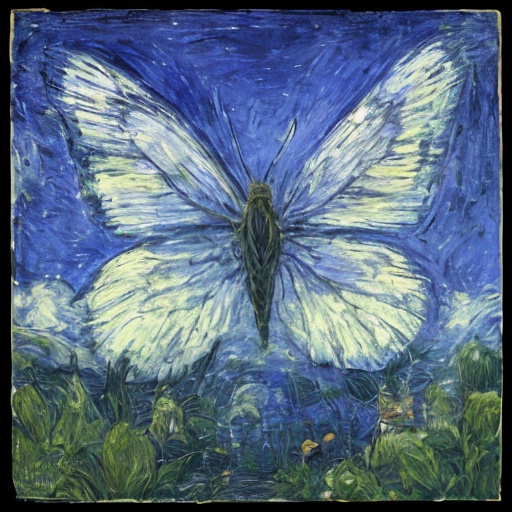}

\\

Style refer & ``Apples'' & ''Butterfly'' 

\end{tabular}
\vspace{-0.3cm}
\caption{Issues of the direct addition method.}
\label{fig_method_2}
\vspace{-0.3cm}
\end{figure}

\noindent\textbf{Direct Addition.} 
To address the issue mentioned above, we can enhance the semantic representation in $\hat{f}^c_l$ by reintroducing the semantic information, as shown below,
\begin{equation}
    \hat{f}^c_l = \lambda * {\rm \mathcal{A}}(Q^c_l, K^s_l, V^s_l) + {\rm \mathcal{A}}(Q^c_l, K^c_l, V^c_l),
\label{equ_method_2}
\end{equation}
where $\lambda \in [0, 1]$ is the weight hyperparameter. We find that this method is less robust and highly dependent on the choice of $\lambda$. For instance, as shown in Fig.~\ref{fig_method_2}, under the same $\lambda$ setting, the generated stylized images in the second row demonstrate a better presentation of the semantics than those in the first row. 

We believe the unstable performance stems from the fact that the two attention operations in Eq.~\ref{equ_method_2} calculate their attention maps {\em separately}. Recalling the process of the standard attention mechanism, the negative numbers are mapped to very small values by the exponentiation function, effectively reducing their contribution to the attention weights matrix. However, in stylized T2I, if the style statistics and semantics information differ significantly, the attention score matrix (before applying the softmax function) between content features $f^c_l$ and representative style statistics $f^s_l$ might contain negative values in some rows. As a result, the exponentiation function cannot work effectively. Therefore, it becomes necessary to introduce an additional weighting coefficient $\lambda$ to balance the two components on the right-hand side of Eq.~\ref{equ_method_2}.

\begin{figure}[!t]
\centering
\begin{tabular}{c@{\hspace{0.3em}}c@{\hspace{0.3em}}c@{\hspace{0.3em}}c@{\hspace{0.3em}}c@{\hspace{0.3em}}c@{\hspace{0.3em}}c}

\includegraphics[width=0.235\linewidth]{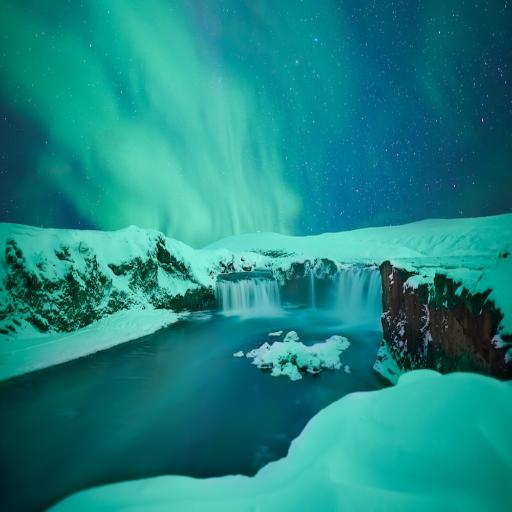} &
\includegraphics[width=0.235\linewidth]{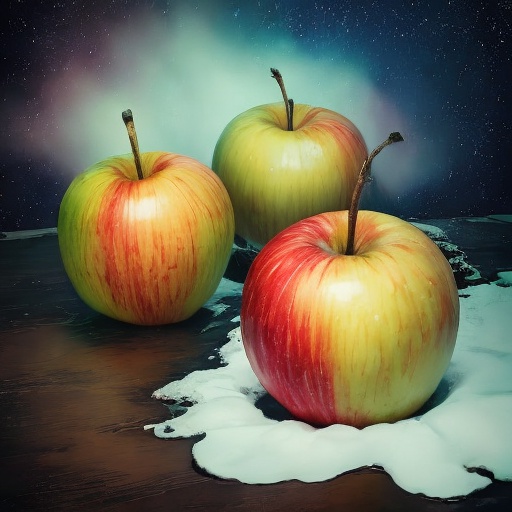} & 
\includegraphics[width=0.235\linewidth]{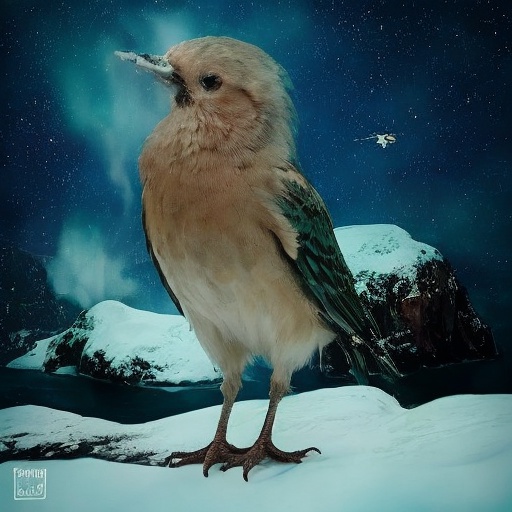} &
\includegraphics[width=0.235\linewidth]{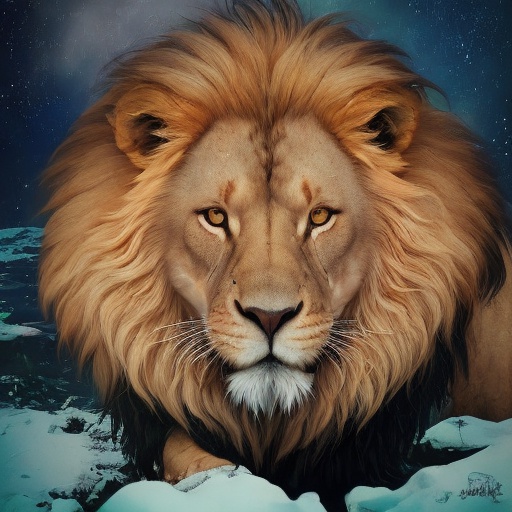}
\\

\includegraphics[width=0.235\linewidth]{images/method_3/mountain.jpg} &
\includegraphics[width=0.235\linewidth]{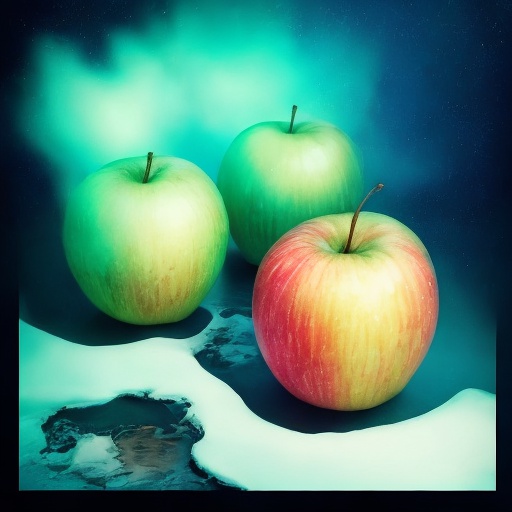} & 
\includegraphics[width=0.235\linewidth]{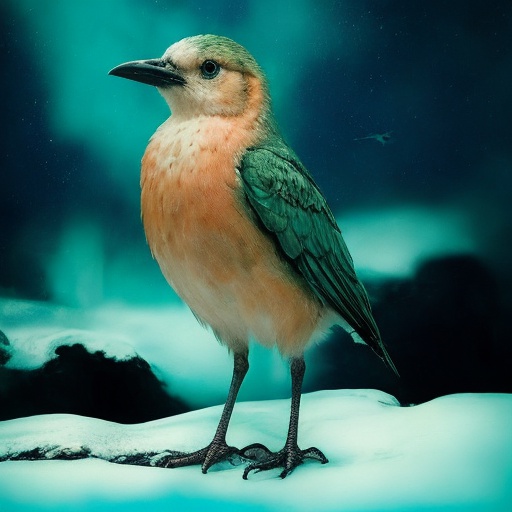} &
\includegraphics[width=0.235\linewidth]{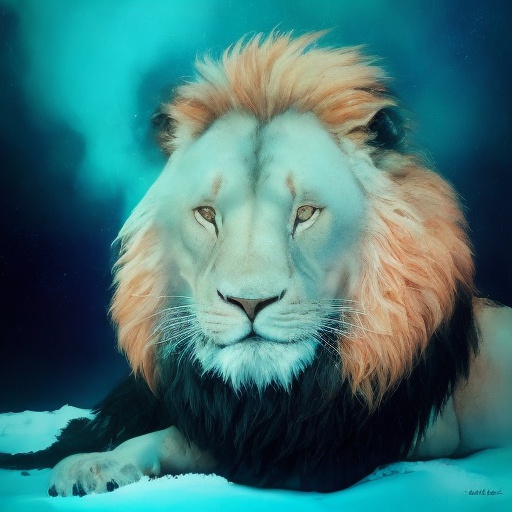}
\\

Style refer & ``Apples'' & ``Bird'' & ``Lion'' \\

\end{tabular}
\vspace{-0.3cm}
\caption{Effect of style distribution normalized.}
\label{fig_method_3_3}
\vspace{-0.3cm}
\end{figure}%

\begin{figure*}[!ht]
\centering
\begin{tabular}{c@{\hspace{0.3em}}c@{\hspace{0.3em}}c@{\hspace{0.3em}}c@{\hspace{0.3em}}c@{\hspace{0.3em}}c@{\hspace{0.3em}}c@{\hspace{0.3em}}}

\includegraphics[width=0.13\linewidth]{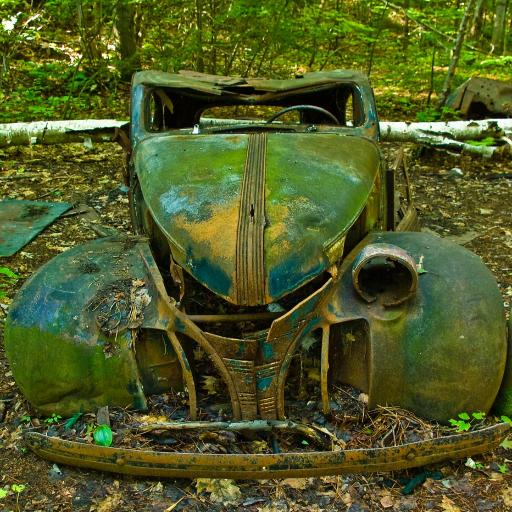} &
\includegraphics[width=0.13\linewidth]{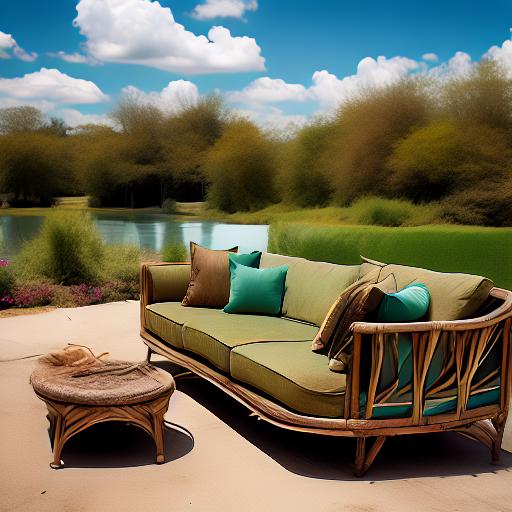} & 
\includegraphics[width=0.13\linewidth]{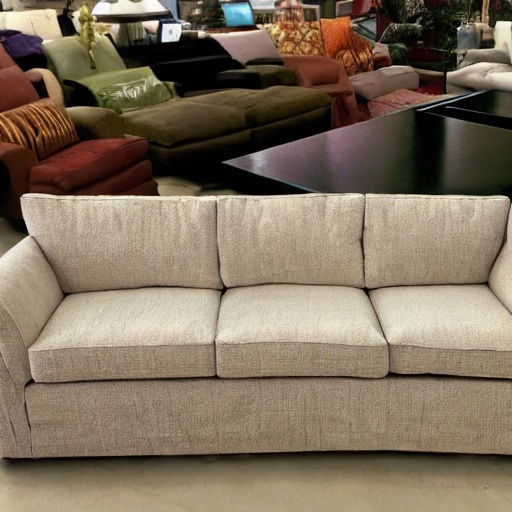} &
\includegraphics[width=0.13\linewidth]{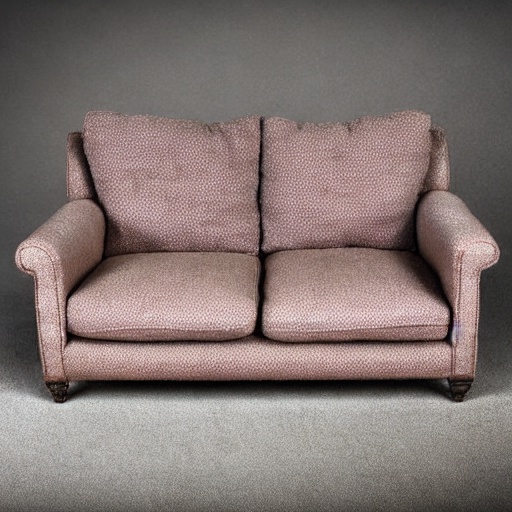} &
\includegraphics[width=0.13\linewidth]{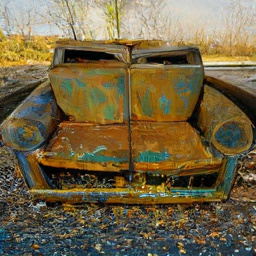} & \includegraphics[width=0.13\linewidth]{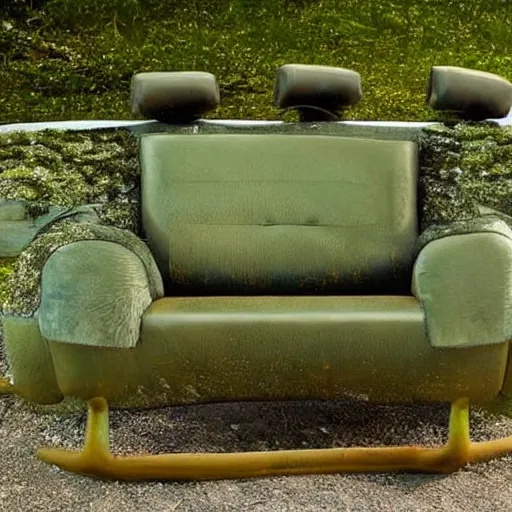} & \includegraphics[width=0.13\linewidth]{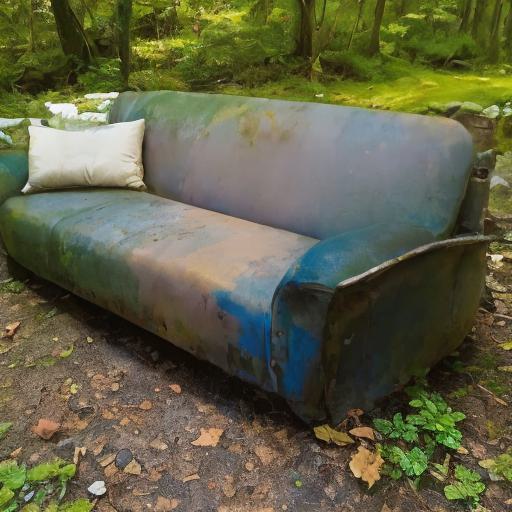}
\\

\includegraphics[width=0.13\linewidth]{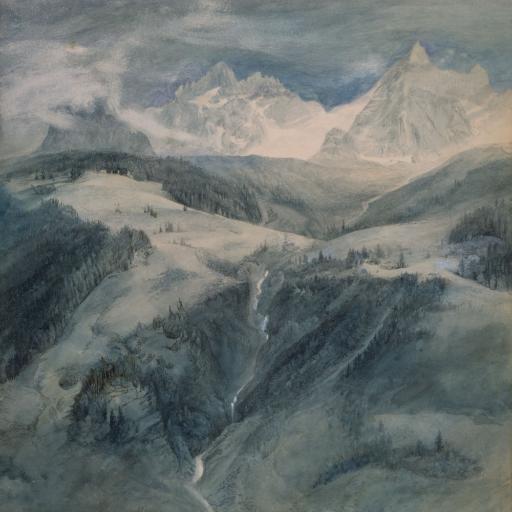} &
\includegraphics[width=0.13\linewidth]{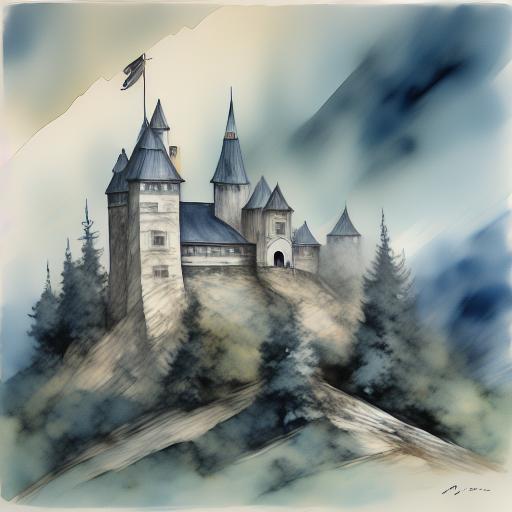} & 
\includegraphics[width=0.13\linewidth]{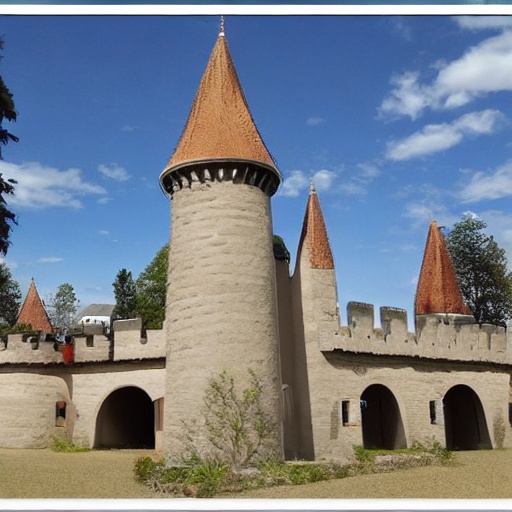} &
\includegraphics[width=0.13\linewidth]{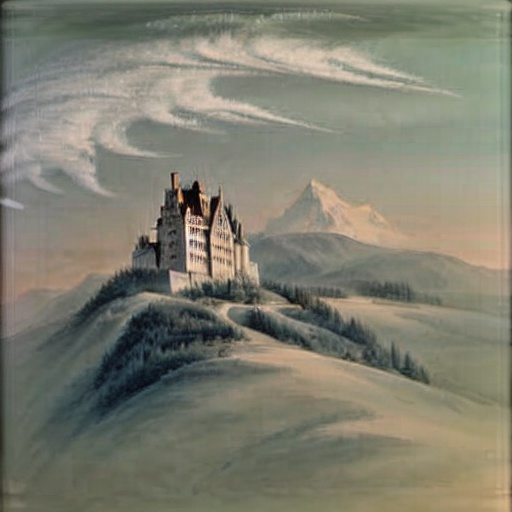} &
\includegraphics[width=0.13\linewidth]{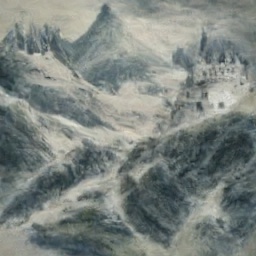} & \includegraphics[width=0.13\linewidth]{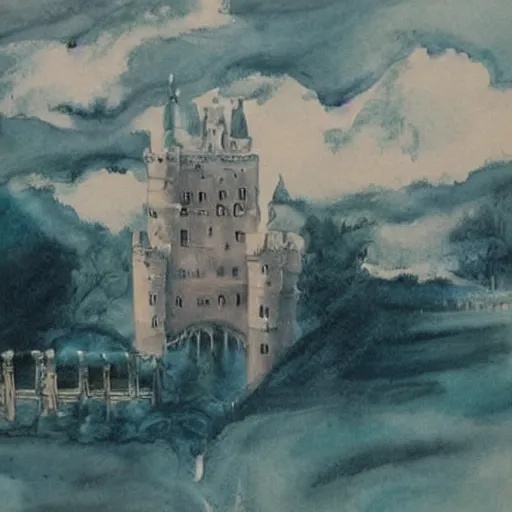} & \includegraphics[width=0.13\linewidth]{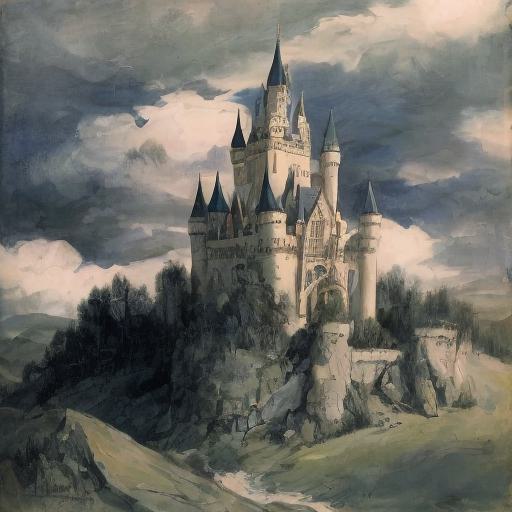}
\\

Style refer & DEADiff & Custom-Diffusion & DreamBooth & StyleDrop & InstaStyle & Ours

\end{tabular}
\vspace{-0.3cm}
\caption{\textbf{Qualitative comparison of personalized T2I generation on various style images.} The prompts for synthesis, listed from top to bottom, are: ``couch" and ``castle".}
\label{fig_comparison}
\vspace{-0.3cm}
\end{figure*}%

\begin{figure*}[!ht]
\centering
\begin{tabular}{c@{\hspace{0.3em}}c@{\hspace{0.3em}}c@{\hspace{0.3em}}c@{\hspace{0.3em}}c@{\hspace{0.3em}}c@{\hspace{0.3em}}c@{\hspace{0.3em}}}

\includegraphics[width=0.13\linewidth]{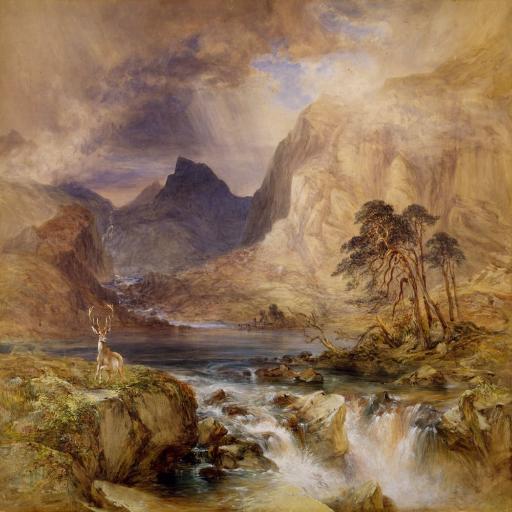} &
\begin{overpic}[width=0.13\linewidth]{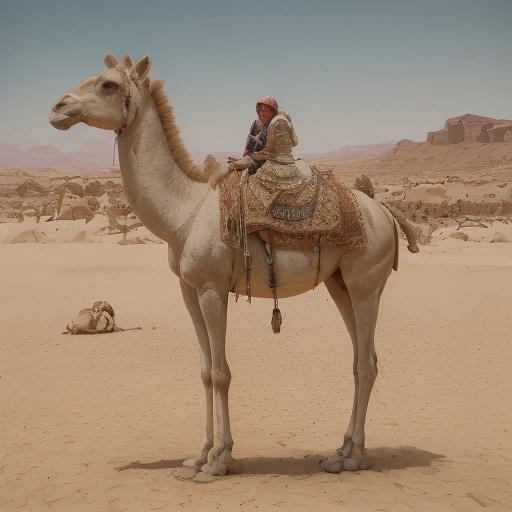}
  \put(10,10){\color{white}\bfseries\itshape\large camel}
\end{overpic} & 
\includegraphics[width=0.13\linewidth]{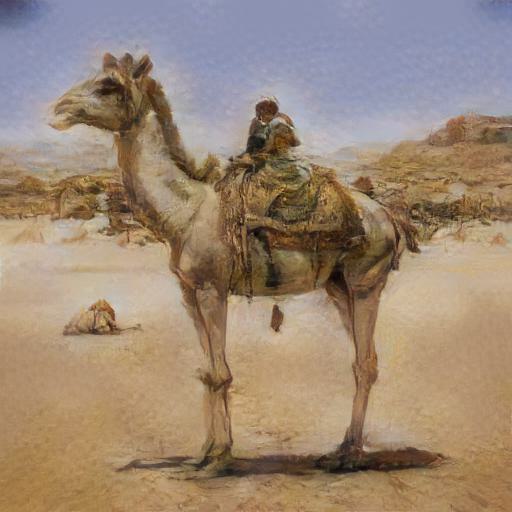} &
\includegraphics[width=0.13\linewidth]{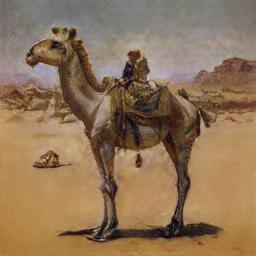} &
\includegraphics[width=0.13\linewidth]{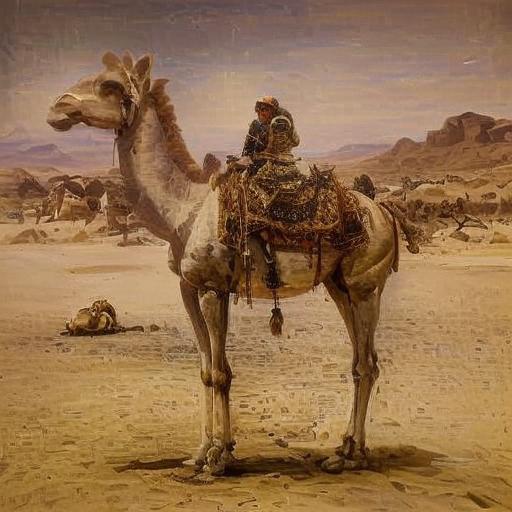} & \includegraphics[width=0.13\linewidth]{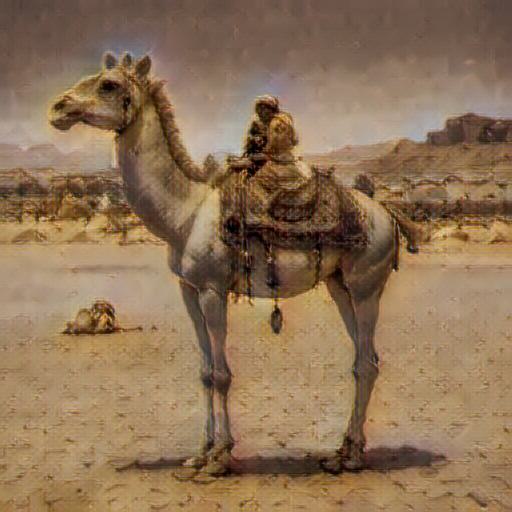} & \includegraphics[width=0.13\linewidth]{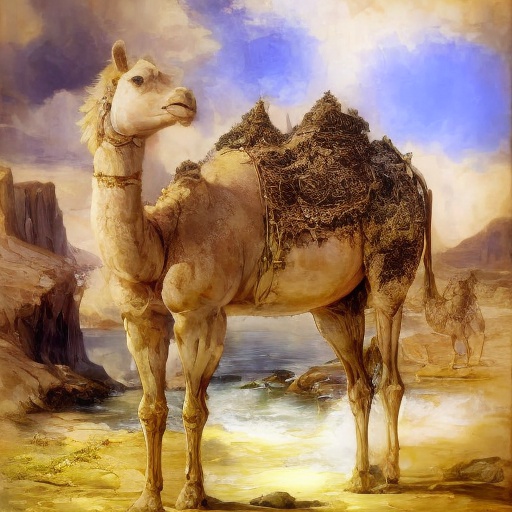}
\\

\includegraphics[width=0.13\linewidth]{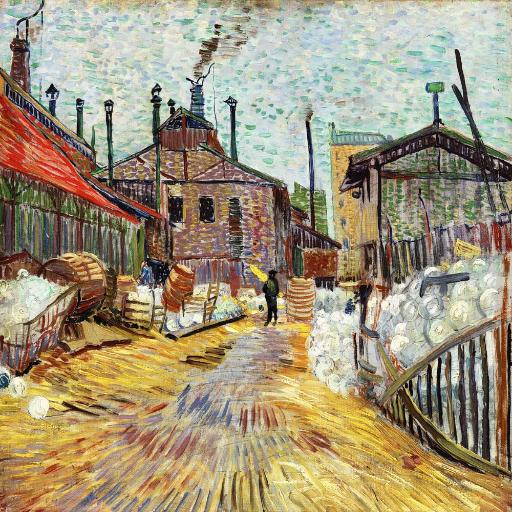} &
\begin{overpic}[width=0.13\linewidth]{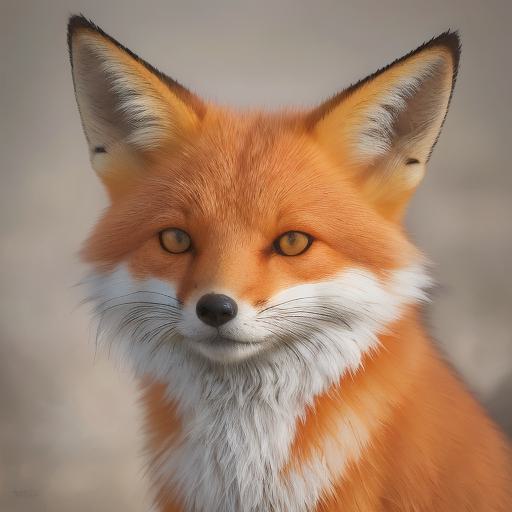}
  \put(10,10){\color{white}\bfseries\itshape\large fox}
\end{overpic} & 
\includegraphics[width=0.13\linewidth]{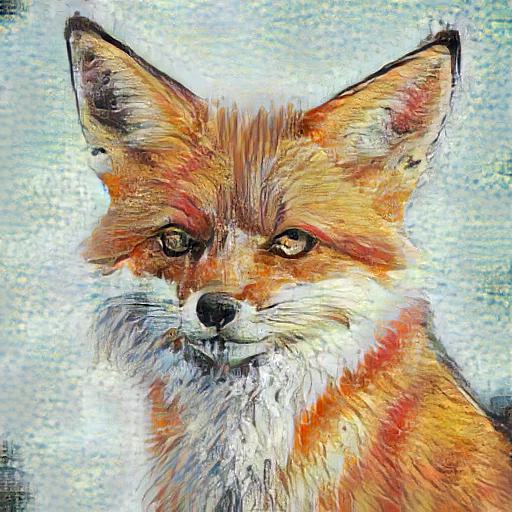} &
\includegraphics[width=0.13\linewidth]{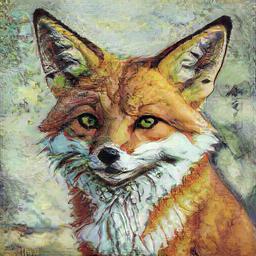} &
\includegraphics[width=0.13\linewidth]{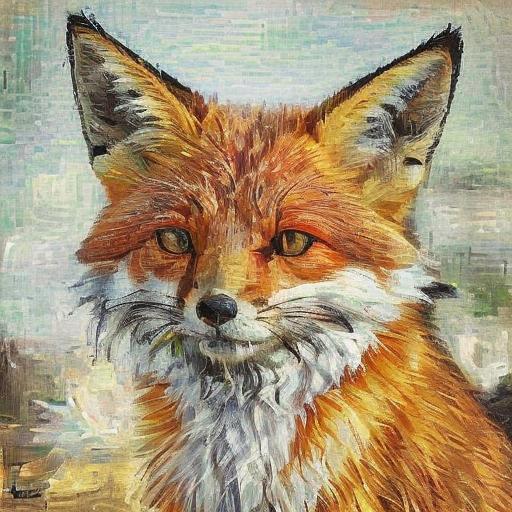} & \includegraphics[width=0.13\linewidth]{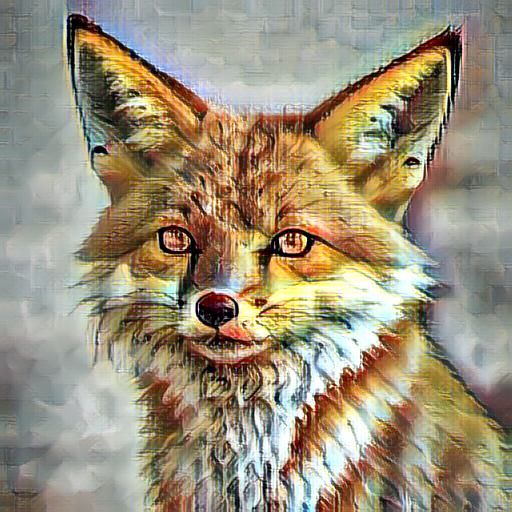} & \includegraphics[width=0.13\linewidth]{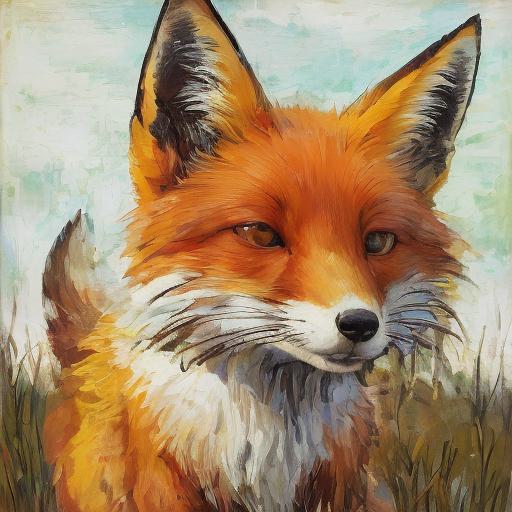}
\\


Style refer & Content & AesPA-Net & CAST & StyleID & CCPL & Ours

\end{tabular}
\vspace{-0.3cm}
\caption{\textbf{Qualitative comparison with style transfer methods.} Content images, displayed in the second column, are used by style transfer methods, whereas our method relies solely on the related prompts shown in white.}
\label{fig_comparison_with_style_transfer}
\vspace{-0.3cm}
\end{figure*}%


\noindent\textbf{Norm Mixture of Self-Attention.} To avoid the aforementioned problem, it is essential to ensure that the calculation of the attention score matrix accounts for both the intra-class semantic differences and the inter-class (semantic and style) information differences, rather than computing these components separately. We first rewrite Eq.~\ref{equ_method_2} in matrix form yielding the following formulation,
\begin{equation}
\begin{aligned}
\hat{f}^c_l = \left[ 
    \lambda \cdot \sigma \left( \frac{Q^c_l {K^s_l}^T}{\sqrt{d}} \right), \ 
    \sigma \left( \frac{Q^c_l {K^c_l}^T}{\sqrt{d}} \right) 
\right] * \left[\begin{array}{l} V^s_l \\ V^c_l \end{array}\right] = M * \hat{V}^T.
\end{aligned}
\label{equ_method_3_1}
\end{equation}
To achieve this uniform attention score matrix calculation, we form 
a new matrix $\left[ 
    \lambda \frac{Q^c_l {K^s_l}^T}{\sqrt{d}}, \frac{Q^c_l {K^c_l}^T}{\sqrt{d}} 
\right]$ by concatenating $
    \lambda ({Q^c_l {K^s_l}^T})/\sqrt{d}$ and 
    $  ({Q^c_l {K^c_l}^T})/{\sqrt{d}}$
and then rewrite $M$ in Eq.~\ref{equ_method_3_1} as $\hat{A}$ as follows,
\begin{equation}
    \hat{M} = \sigma \left( \left[ 
    \lambda \frac{Q^c_l {K^s_l}^T}{\sqrt{d}}, \frac{Q^c_l {K^c_l}^T}{\sqrt{d}} 
\right] \right).
\label{equ_method_3_2}
\end{equation}

Eq.~\ref{equ_method_3_2} performs the softmax operation on inter-class and intra-class information as a whole and avoids the potential failures that arise from using the exponentiation function to project the attention score matrix into the attention weights matrix as mentioned above. The stylized content features $\hat{f}^c_l$ can be obtained through,
\begin{equation}
\hat{f}^c_l = \sigma \left( \left[ 
    \lambda \frac{Q^c_l {K^s_l}^T}{\sqrt{d}}, \frac{Q^c_l {K^c_l}^T}{\sqrt{d}} 
\right] \right) * \left[\begin{array}{l} V^s_l \\ V^c_l \end{array}\right],
\label{equ_method_3_3}
\end{equation}
which enables the model to more adaptively aggregate semantic and style information, while reducing its heavy dependence on the choice of $\lambda$. 

Eq.~\ref{equ_method_3_3} leverages each input patch of the content features $f^c_l$ to query the most relevant information from the representative style statistics $f^s_l$. However, this approach can sometimes lead to a mismatch in the style distribution between $\hat{f}^c_l$ and $f^s_l$, as it does not account for the global style distribution. As illustrated in the first row in Fig.~\ref{fig_method_3_3}, this mismatch manifests as global color inconsistency between the generated stylized image and the reference style images.

We thus introduce the style distribution normalized process before applying Eq.~\ref{equ_method_3_3} inspired by AdaIN~\cite{huang2017arbitrary}. This process can be mathematically expressed as follows,
\begin{equation}
    f_l^c = \delta_l^s * \left( \frac{f_l^c - \mu_l^c}{\delta_l^c} \right) + \mu_l^s,
\label{eq_adain}
\end{equation}
where $\mu_l^s$, $\mu_l^c$ $\delta_l^s$, and $\delta_l^c$ represent the mean and standard deviation of content feature $f^c_l$ and representative style statistics $f^s_l$, respectively. As shown in Fig.~\ref{fig_method_3_3} (lower row), after applying the style distribution normalization, the global color tone becomes more consistent with the style image compared to the case without this normalization (upper row). Finally, we collectively refer to the above two processes as the \textit{norm mixture of self-attention}.
\section{Experiments}

\subsection{Evaluation Benchmark and Metrics.} Following the previous methods~\cite{cui2025instastyle, sohn2023styledrop, everaert2023diffusion}, we utilize a dataset of 60 style images as the evaluation benchmark. The generated stylized images are assessed quantitatively from two perspectives. First, we compute the CLIP score between the generated stylized images and the style images to evaluate the style consistency. Second, we calculate the CLIP score between the generated stylized images and the given prompt to assess the content fidelity~\cite{radford2021learning,cui2025instastyle}.

\subsection{Stylized Image Synthesis}
We compare our method with recent methods using their open-source codes: Textual Inversion~\cite{gal2022image}, Custom Diffusion~\cite{kumari2023multi}, DreamBooth~\cite{ruiz2023dreambooth}, StyleDrop~\cite{sohn2023styledrop}, DEADiff~\cite{qi2024deadiff}, and InstaStyle~\cite{cui2025instastyle}. Additionally, we evaluate our method against style transfer methods, including AdaAttn~\cite{liu2021adaattn}, CCPL~\cite{wu2022ccpl}, StyTr$^2$~\cite{deng2022stytr2}, CAST~\cite{zhang2022domain}, AesPA-Net~\cite{hong2023aespa}, and StyleID~\cite{chung2024style}, to demonstrate the superiority of our method. Notably, style transfer methods require a content image, while our method requires only text prompts. To facilitate the comparison, we generate the content images from the same text prompts and feed them to the style transfer methods. While this enables the comparison, it is not entirely fair to our method, as the content images fed to style transfer methods may provide additional content and semantic information that text prompts alone cannot convey.




\noindent\textbf{Qualitative Results.} As shown in Fig.~\ref{fig_comparison} and Fig.~\ref{fig_additional_comparison}, Custom Diffusion struggles to capture style patterns effectively, while DreamBooth and StyleDrop perform slightly better. Their limited performance is mainly due to the availability of only a single style image for fine-tuning in our scenario. The improved performance of DEADiff can be attributed to its well-designed training dataset. InstaStyle achieves better results by leveraging DDIM Inversion to generate multiple images with a similar style pattern, but it is prone to structural inaccuracies. Benefiting from representative style statistics and NMSA, our method generates stylized images with fine-grained details and higher fidelity without fine-tuning. Fig.~\ref{fig_comparison_with_style_transfer} and Fig.~\ref{fig_additiaonl_comparison_with_style_transfer} further compare our method with style transfer approaches. Our method preserves style details effectively while maintaining comparable content fidelity.


\noindent\textbf{Quantitative Results.} Following the evaluation setting of the previous works~\cite{cui2025instastyle,sohn2023styledrop}, we use 100 objects from CIFAR100~\cite{krizhevsky2009learning} as input prompts for models. The generated images from different models are evaluated based on two aspects.
As shown in Tab.~\ref{tab_quantitative_comparisons}, our method achieves the highest style consistency while maintaining comparable content fidelity, demonstrating its ability to generate images that align with the style image while accurately following the given prompt. Compared to style transfer methods like AdaAttn, CCPL, and CAST, which use content images, our text-based approach still achieves a comparable content score (28.24) and significantly higher style consistency.

\begin{table}[!t]
\centering
\begin{tabular}{ccc}
\hline
Method            & Style $\uparrow$ & Content $\uparrow$ \\ \hline
AdaAttn (ST)~\cite{liu2021adaattn}           & 58.7 & 28.70   \\
CCPL (ST)~\cite{wu2022ccpl}              & 54.2 & 28.60   \\
StyTr$^2$ (ST)~\cite{deng2022stytr2}            & 56.0 & 28.20   \\
CAST (ST)~\cite{zhang2022domain}              & 59.4 & 28.50   \\
AesPA-Net (ST)~\cite{hong2023aespa}         & 60.4 & 28.40   \\
Textual Inversion~\cite{gal2022image} & 56.2 & 27.80   \\
Custom Diffusion~\cite{kumari2023multi}  & 61.1 & 28.00   \\
DreamBooth~\cite{ruiz2023dreambooth}        & 56.1 & 23.54   \\
StyleDrop~\cite{sohn2023styledrop}         & 60.2 & 23.58   \\
InstaStyle~\cite{cui2025instastyle}        & 65.5 & \textbf{29.40}   \\ \hline
Ours              & \textbf{67.3} & 28.24   \\ \hline
\end{tabular}
\caption{\textbf{Quantitative comparisons between the baselines and our method.} Note that ``ST'' means the method designed for the style transfer method.}
\label{tab_quantitative_comparisons}
\vspace{-0.9cm}
\end{table}

\noindent\textbf{User Study.} Following prior works~\cite{cui2025instastyle,deng2022stytr2,sohn2023styledrop}, we conduct a user study to align with human evaluation standards. Ten independent evaluators were presented with anonymized image pairs, each containing one result from our method and one from a baseline, displayed in random order. They were instructed to prioritize stylistic quality and semantic content preservation. As shown in Tab.~\ref{tab_user_study}, our method receives a higher preference. 


\begin{table}[!ht]
\centering
\begin{tabular}{cccc}
\hline
Method & Good & Par & Ours  \\ \hline
AdaAttn (ST)~\cite{liu2021adaattn} & 0.28 & 0.18 & 0.54   \\
CCPL (ST)~\cite{wu2022ccpl} & 0.36 & 0.20 & 0.44   \\
StyTr$^2$ (ST)~\cite{deng2022stytr2} & 0.28 & 0.24 & 0.48   \\
CAST (ST)~\cite{zhang2022domain} & 0.36 & 0.12 & 0.52   \\
AesPA-Net (ST)~\cite{hong2023aespa} & 0.16 & 0.16 & 0.68   \\
Textual Inversion~\cite{gal2022image} & 0.20 & 0.07 & 0.73   \\
Custom Diffusion~\cite{kumari2023multi} & 0.08 & 0.10 & 0.82   \\
DreamBooth~\cite{ruiz2023dreambooth} & 0.16 & 0.20 & 0.64   \\
StyleDrop~\cite{sohn2023styledrop} & 0.34 & 0.16 & 0.50   \\
InstaStyle~\cite{cui2025instastyle} & 0.31 & 0.23 & 0.46   \\ \hline
\end{tabular}
\caption{\textbf{User study.} The results illustrate the proportion of votes showing whether the comparison method is favored over ours, is on par with ours, or falls short of ours.}
\label{tab_user_study}
\vspace{-0.5cm}
\end{table}

\subsection{Analysis} 
In this section, we conduct the ablation study to analyze the effectiveness of each component.

\noindent\textbf{Impact of Inference Step.} As shown in Fig.~\ref{fig_inference_number_steps}, considering the trade-off, we ultimately set the number of sampling steps to 6.

\noindent\textbf{Impact of Timestep.} In Sec.~\ref{sec:extracting_representative_style_statistics}, we describe first adding noise to $z^s$ and then using LCMs to extract style statistics. We conduct an ablation study to investigate the impact of the timestep on the performance. As shown in Tab.~\ref{table_timestep_of_style-consistency_features}, the content fidelity remains relatively stable, whereas the style consistency scores at $t=0$ and $t=500$ are notably low. We select a timestep of $t=200$ for extracting style statistics, as it yields a higher content fidelity score.

\begin{table}[!ht]
\centering
\scalebox{0.9}{
\begin{tabular}{ccccccc}
\hline
Timestep & 0 & 100 & 200 & 300 & 400 & 500 \\ \hline
Style $\uparrow$ & 65.17 & 67.30 & 67.29 & 66.94 & 66.29 & 64.95 \\
Content $\uparrow$ & 28.08 & 28.20 & 28.24 & 28.32 & 28.30 & 28.37 \\
\hline
\end{tabular}
}
\caption{\textbf{Quantitative comparison of different timestep.} $\uparrow$ means the higher the better.}
\label{table_timestep_of_style-consistency_features}
\vspace{-0.5cm}
\end{table}

\begin{figure}[!ht]
\centering
\begin{tabular}{c@{\hspace{0.3em}}c@{\hspace{0.3em}}c@{\hspace{0.3em}}c@{\hspace{0.3em}}c@{\hspace{0.3em}}c@{\hspace{0.3em}}c}

\includegraphics[width=0.18\linewidth]{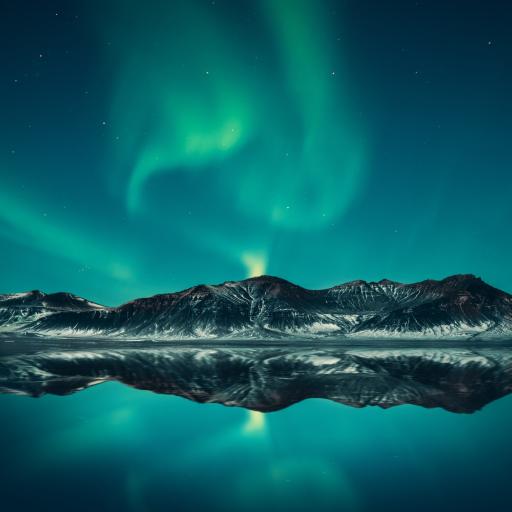} &
\includegraphics[width=0.18\linewidth]{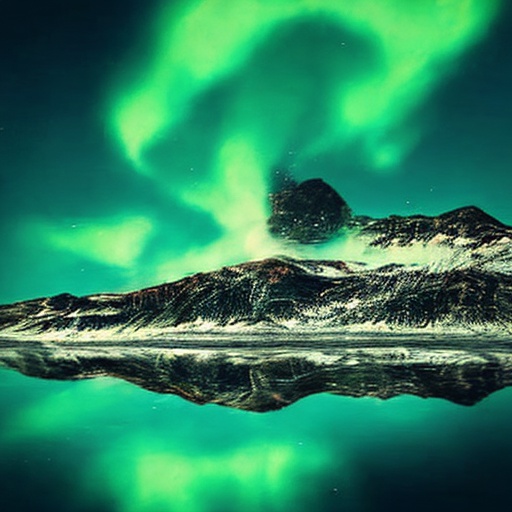} & 
\includegraphics[width=0.18\linewidth]{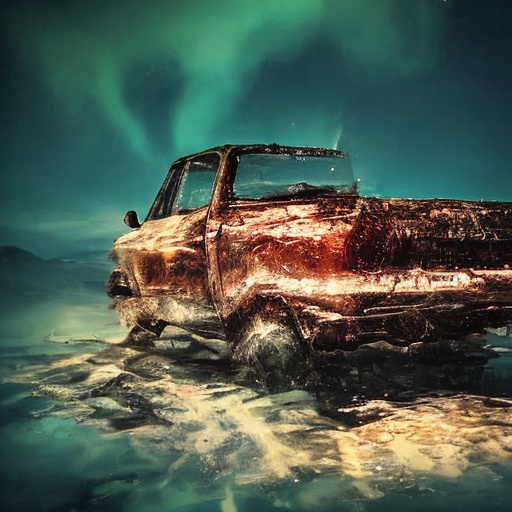} &
\includegraphics[width=0.18\linewidth]{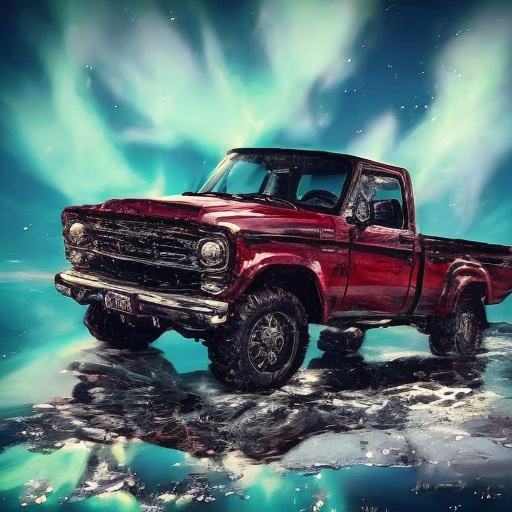} &
\includegraphics[width=0.18\linewidth]{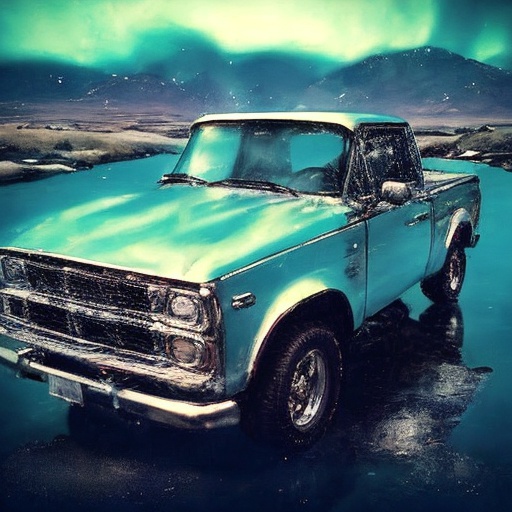}
\\

{\scriptsize Style refer} & {\scriptsize Eq.~\ref{eq_method_1}} & {\scriptsize Eq.~\ref{equ_method_2}} & {\scriptsize Eq.~\ref{equ_method_3_3}} & {\scriptsize Eq.~\ref{equ_method_3_3} and~\ref{eq_adain}} \\

\end{tabular}
\vspace{-0.3cm}
\caption{\textbf{Effects of different attention controls.} Here, the operations represented by different formulas are shown in Tab.~\ref{table_quantitative_comparison_of_different_attention_controls}.}
\label{fig_analysis_different_ac}
\vspace{-0.3cm}
\end{figure}

\noindent\textbf{Impact of Attention Controls.} To evaluate the effectiveness of the different attention controls described in Sec.~\ref{sec_mixture_of_style_distribution_normalized_self-attention}, we provide a visual comparison in Fig.~\ref{fig_analysis_different_ac}. and quantitative results in Tab.~\ref{table_quantitative_comparison_of_different_attention_controls}. Direct replacing and direct adding achieve relatively high style consistency scores but significantly lower content fidelity, indicating a tendency to capture style statistics at the expense of semantic content details from prompts. The visual results in Fig.~\ref{fig_analysis_different_ac} further support this conclusion. In contrast, both MSA and NMSA offer a better balance between style and content. Notably, the latter achieves higher style consistency with only a 0.23 drop in content fidelity, producing images that more closely resemble the style image in overall color as shown in Fig.~\ref{fig_analysis_different_ac}.

\begin{table}[!ht]
\centering
\begin{tabular}{ccc}
\hline
Operations & Style $\uparrow$ & Content $\uparrow$ \\ \hline
Direct replacing (Eq.~\ref{eq_method_1}) & 86.79  & 19.67 \\
Direct addition (Eq.~\ref{equ_method_2})  & 76.04 & 24.50  \\
Mixture of self-attention (Eq.~\ref{equ_method_3_3}) & 65.46 & 28.43  \\
NMSA (Eq.~\ref{equ_method_3_3} and Eq.~\ref{eq_adain}) & 67.30 & 28.20 \\ \hline
\end{tabular}
\caption{\textbf{Quantitative comparison of different attention controls.} Here, NMSA represents the norm mixture of self-attention.}
\label{table_quantitative_comparison_of_different_attention_controls}
\vspace{-0.8cm}
\end{table}
\section{Conclusion}
In this paper, we present OmniPainter, a novel and efficient stylized text-to-image generation method that eliminates the need for inversion. Our approach begins by extracting representative style statistics from reference style images. To seamlessly incorporate these statistics into image generation, we propose a norm mixture of self-attention. Experimental results demonstrate the effectiveness and superiority of our method compared to existing approaches.

\bibliographystyle{ACM-Reference-Format}
\bibliography{sample-bibliography}

\clearpage
\appendix
\section{Performance and Efficiency Analysis}
\begin{figure}[!ht]
\centering
\includegraphics[width=0.85\linewidth]{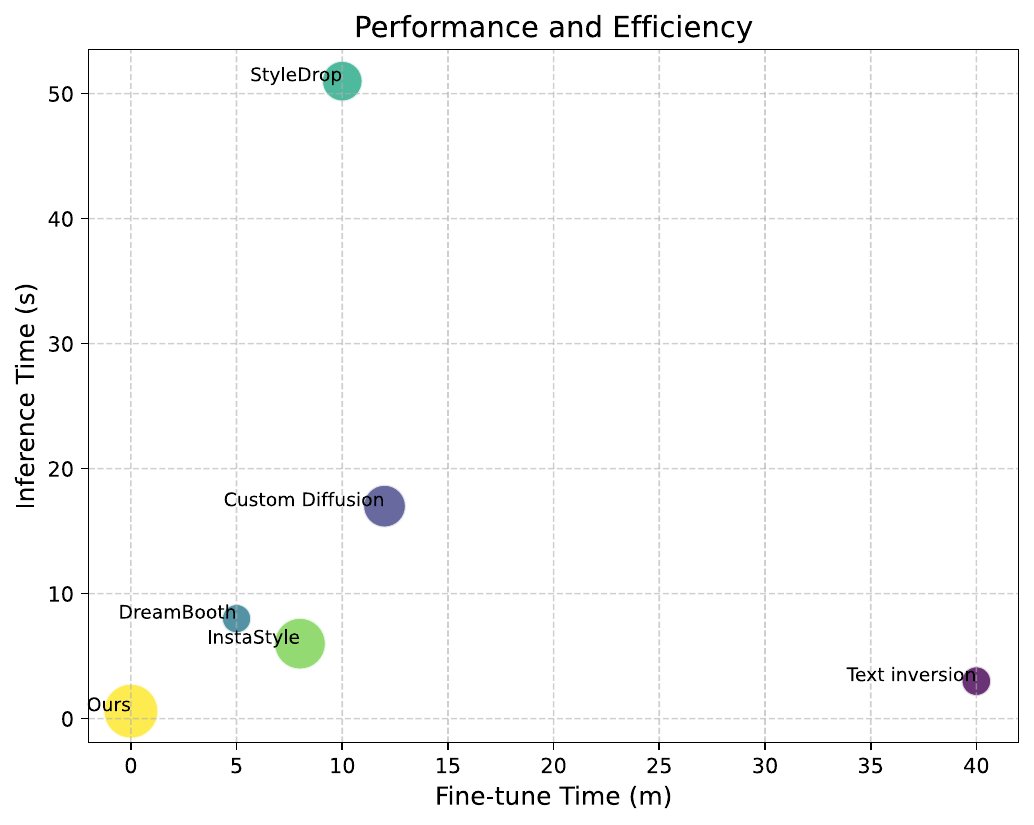}
\caption{\textbf{The comparison of performance and efficiency.} Our method delivers exceptional performance results without the need for fine-tuning and achieves the shortest inference time.}
\label{fig_performance_vs_efficiency}
\end{figure}

As shown in Fig.~\ref{fig_performance_vs_efficiency}, we compare our method with five diffusion-based methods, including Textual Inversion~\cite{gal2022image}, Custom Diffusion~\cite{kumari2023multi}, DreamBooth~\cite{ruiz2023dreambooth}, StyleDrop~\cite{sohn2023styledrop}, and InstaStyle~\cite{cui2025instastyle}, in terms of fine-tuning time, inference time, and their corresponding performance. In this figure, the size of the colored circle represents performance: the larger the circle, the better the performance. Generally, methods closer to the bottom-left corner with larger circles indicate better overall performance. Our method achieves exceptional results without requiring fine-tuning and delivers the shortest inference time.

\section{Impact of Inference Number Steps}
\begin{figure}[!ht]
\centering
\includegraphics[width=0.85\linewidth]{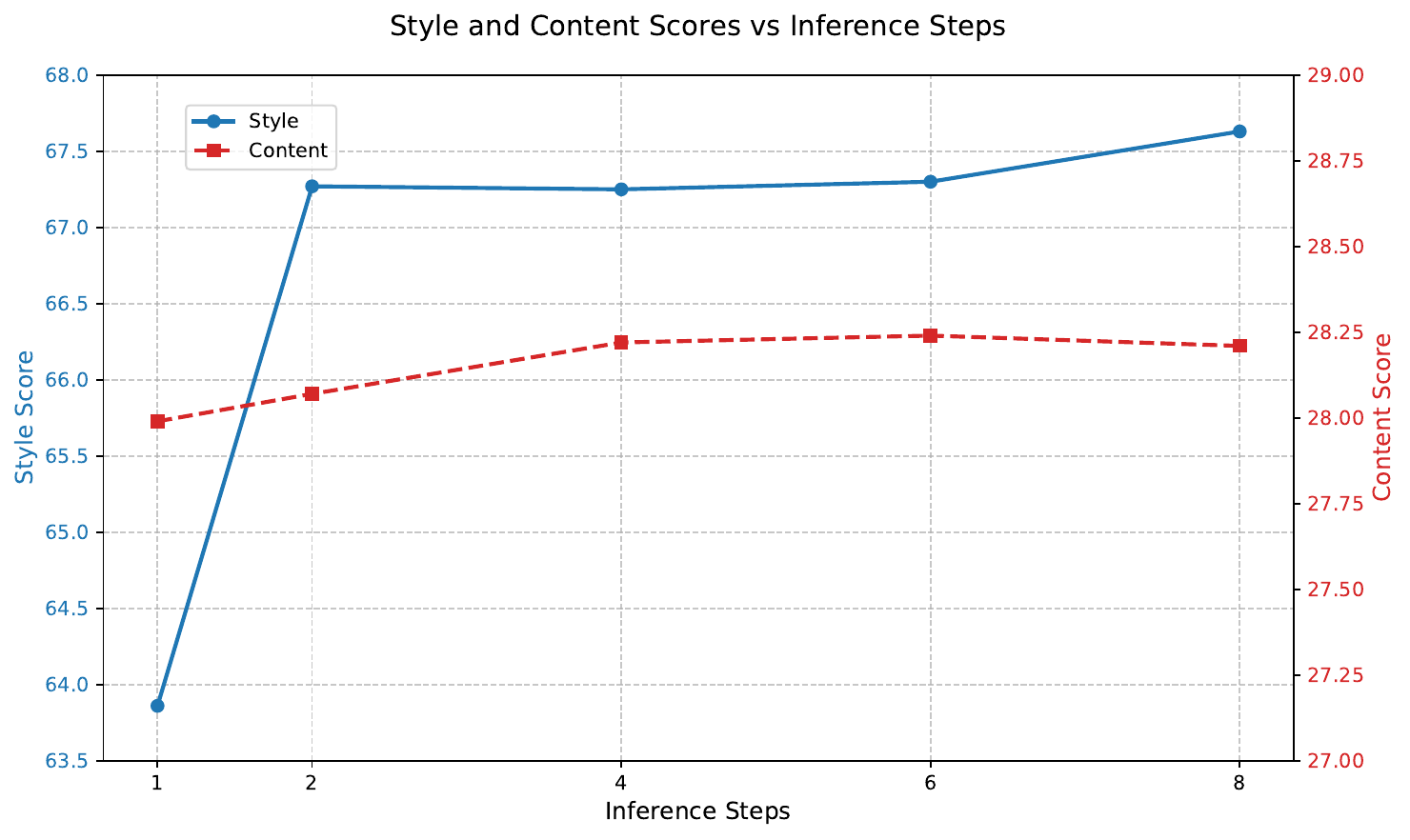}
\caption{Quantitative comparison of different inference number steps.}
\label{fig_inference_number_steps}
\end{figure}

Fig.~\ref{fig_inference_number_steps} presents the quantitative results of our method for image generation at different sampling steps. As shown in this figure, increasing the number of sampling steps during inference can generally enhance both style similarity scores and content fidelity. Considering the trade-off between inference time and performance, we ultimately set the number of sampling steps to 6.

\section{Visualization of One Timestep Denoising}
\begin{figure}[!ht]
\centering
\begin{tabular}{c@{\hspace{0.3em}}c@{\hspace{0.3em}}c@{\hspace{0.3em}}c@{\hspace{0.3em}}c@{\hspace{0.3em}}c@{\hspace{0.3em}}c}

\includegraphics[width=0.18\linewidth]{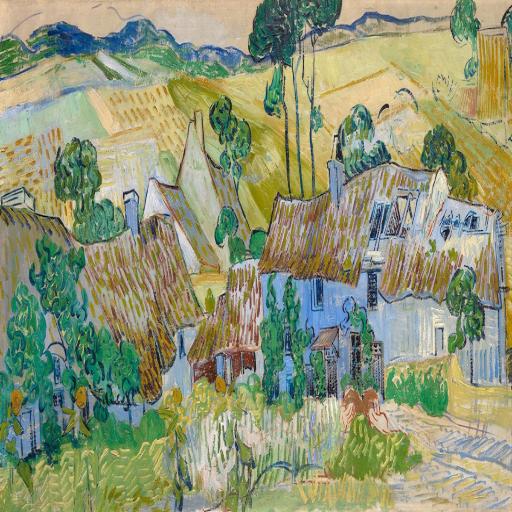} &
\includegraphics[width=0.18\linewidth]{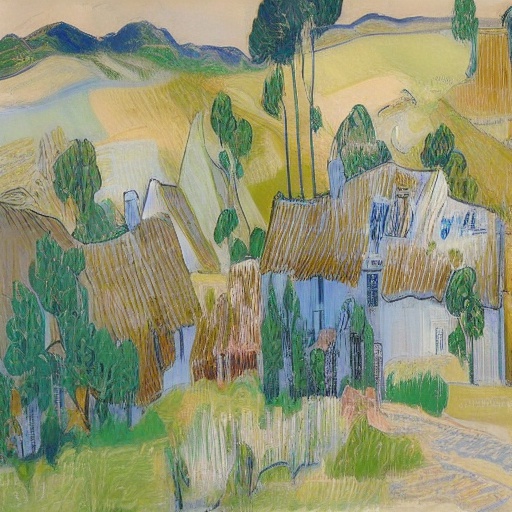} & 
\includegraphics[width=0.18\linewidth]{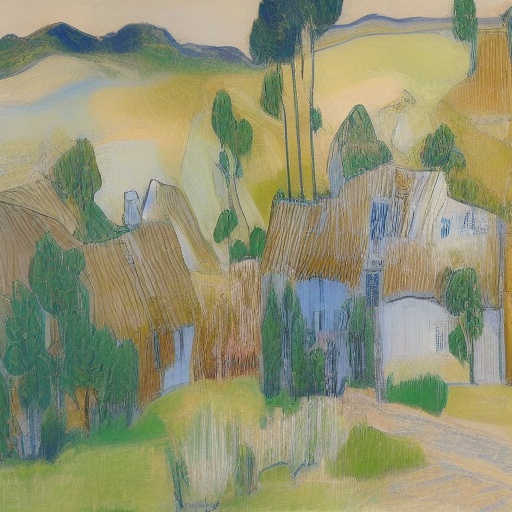} &
\includegraphics[width=0.18\linewidth]{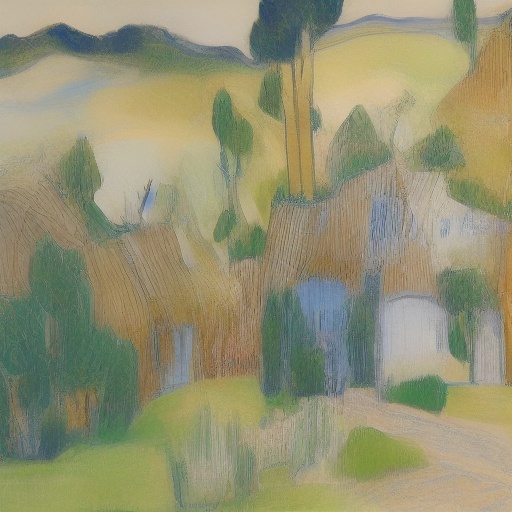} &
\includegraphics[width=0.18\linewidth]{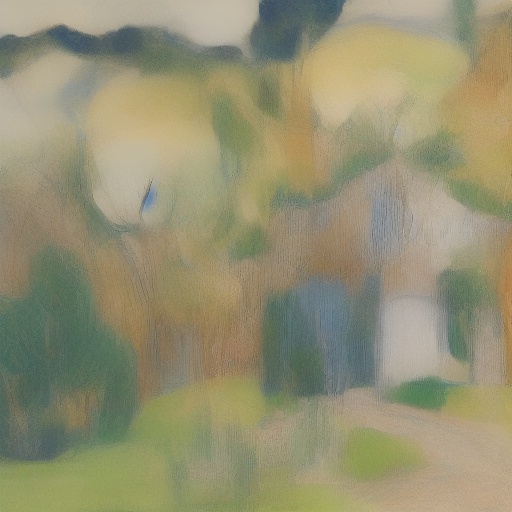}
\\

\includegraphics[width=0.18\linewidth]{images/analysis_one_timestep_denoise/5.jpg} &
\includegraphics[width=0.18\linewidth]{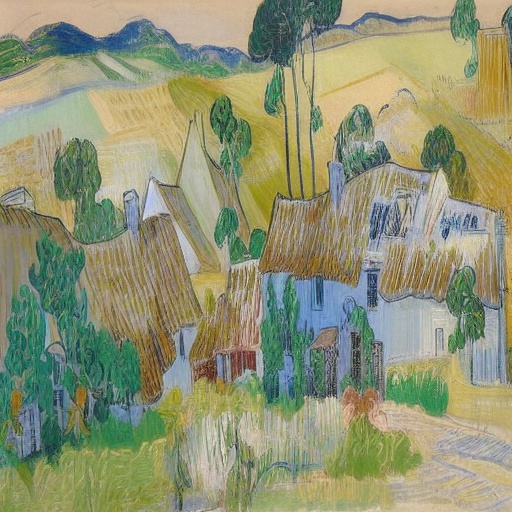} & 
\includegraphics[width=0.18\linewidth]{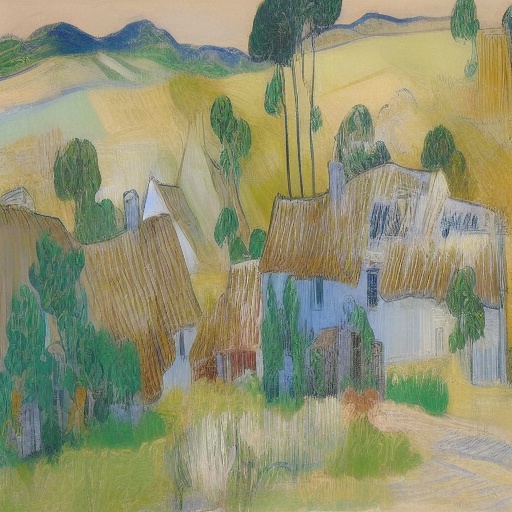} &
\includegraphics[width=0.18\linewidth]{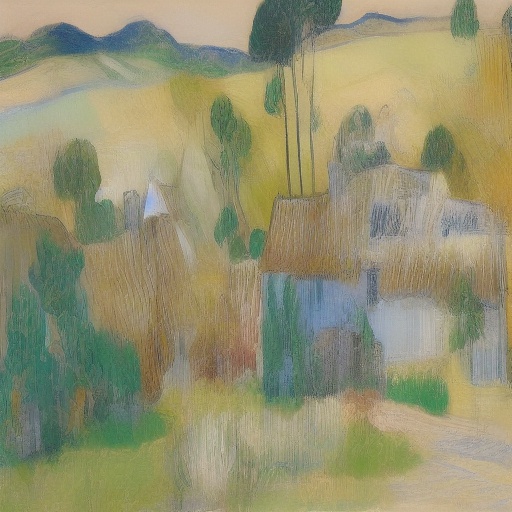} &
\includegraphics[width=0.18\linewidth]{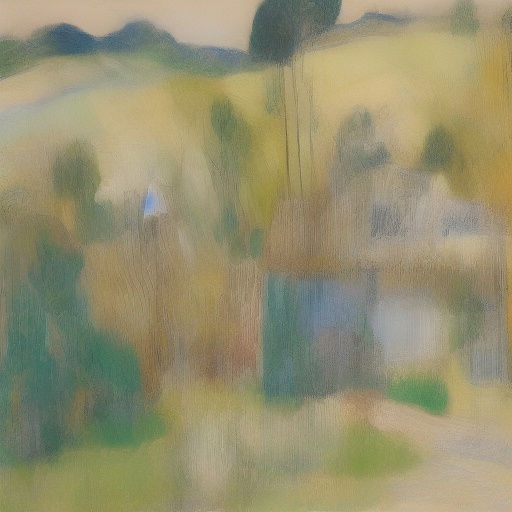}
\\

\includegraphics[width=0.18\linewidth]{images/analysis_one_timestep_denoise/5.jpg} &
\includegraphics[width=0.18\linewidth]{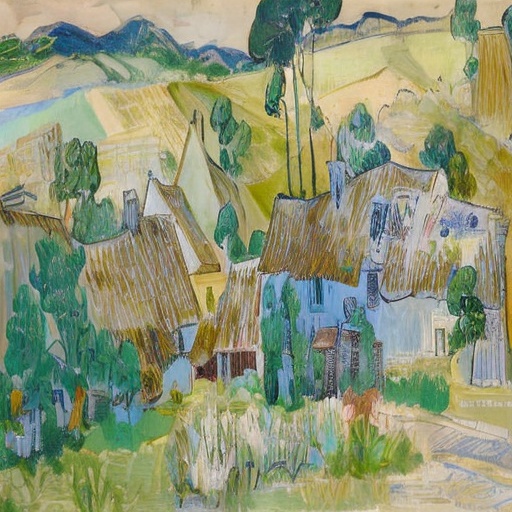} & 
\includegraphics[width=0.18\linewidth]{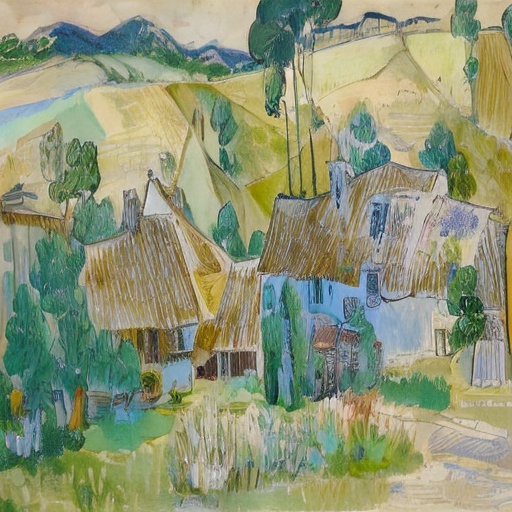} &
\includegraphics[width=0.18\linewidth]{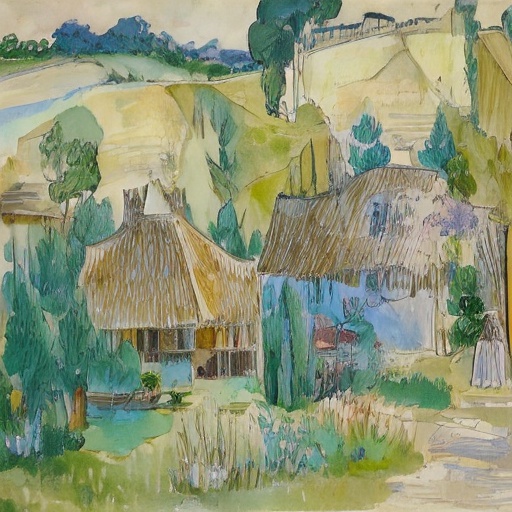} &
\includegraphics[width=0.18\linewidth]{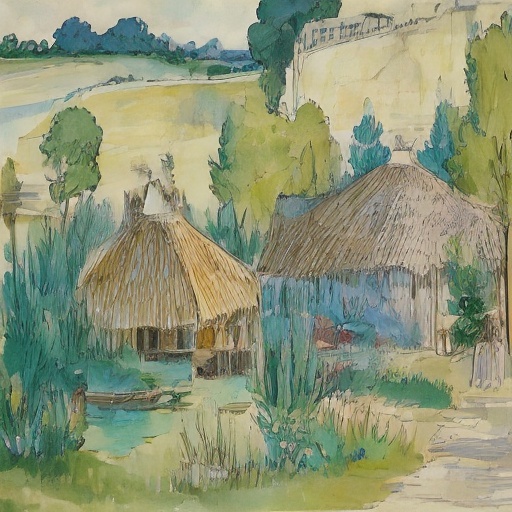}
\\

{\scriptsize Style image} & {\scriptsize t = 100} & {\scriptsize t = 200} & {\scriptsize t = 300} & {\scriptsize t = 400} \\

\end{tabular}
\caption{\textbf{Visualization of denoised style images by using one timestep and three different combinations.} The denoised images in the first, second, and third rows are obtained using Eq.\ref{equ_diffusion_forward} combined with SD, DDIM Inversion combined with SD, and Eq.\ref{equ_diffusion_forward} combined with LCMs, respectively.}
\label{fig_vis_one_timestep_denosing}
\end{figure}

In Sec.~\ref{sec:extracting_representative_style_statistics}, we demonstrate that the combination of Eq.~\ref{equ_diffusion_forward} + LCMs can also effectively extract the representative style statistics from noisy style images. Here, we present the visualizations of one timestep denoised results by using three different combinations. In Fig.\ref{fig_one_timestep_denosing_with_different_methods}, we observe that the CLIP feature similarities between DDIM Inversion combined with SD and Eq.\ref{equ_diffusion_forward} combined with LCMs are quite similar. However, in Fig.\ref{fig_vis_one_timestep_denosing}, the denoised style images obtained using Eq.\ref{equ_diffusion_forward} combined with SD and DDIM Inversion combined with SD appear more blurry compared to those generated by Eq.\ref{equ_diffusion_forward} combined with LCMs. The comparison between these two figures further demonstrates that LCMs can effectively extract representative style statistics from reference style images.

\section{Visual Comparison analysis}
\begin{figure*}[!ht]
\centering
\begin{tabular}{c@{\hspace{0.3em}}c@{\hspace{0.3em}}c@{\hspace{0.3em}}c@{\hspace{0.3em}}c@{\hspace{0.3em}}c@{\hspace{0.3em}}c@{\hspace{0.3em}}}

\includegraphics[width=0.13\linewidth]{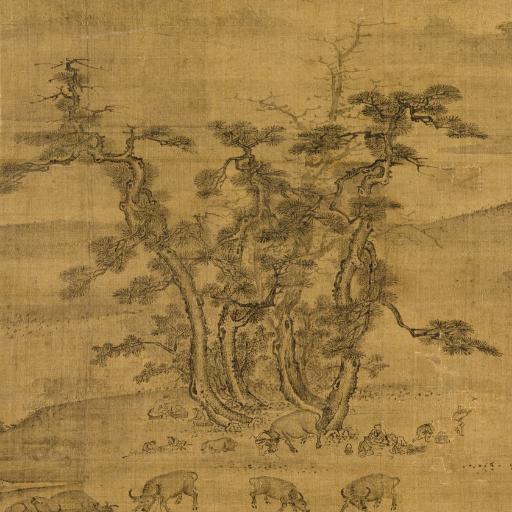} &
\includegraphics[width=0.13\linewidth]{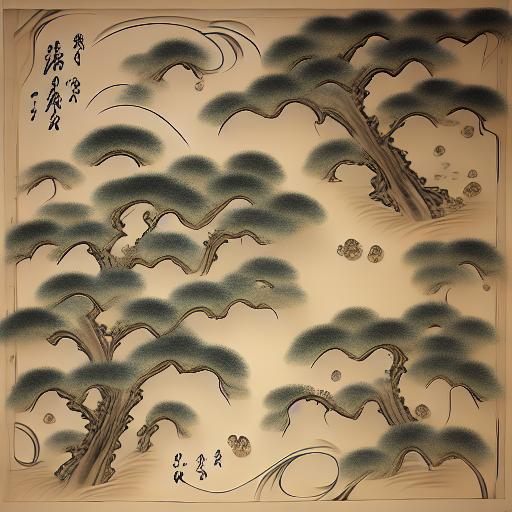} & 
\includegraphics[width=0.13\linewidth]{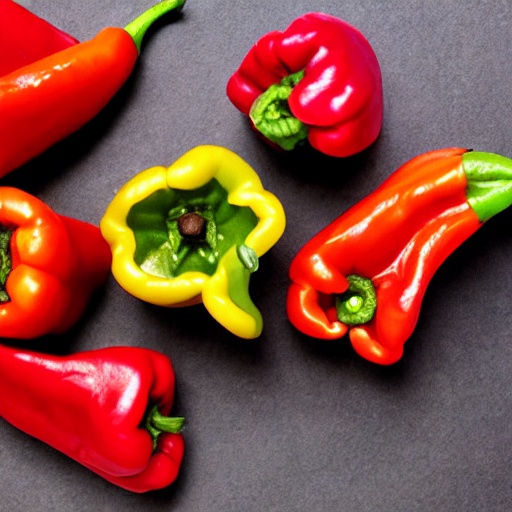} &
\includegraphics[width=0.13\linewidth]{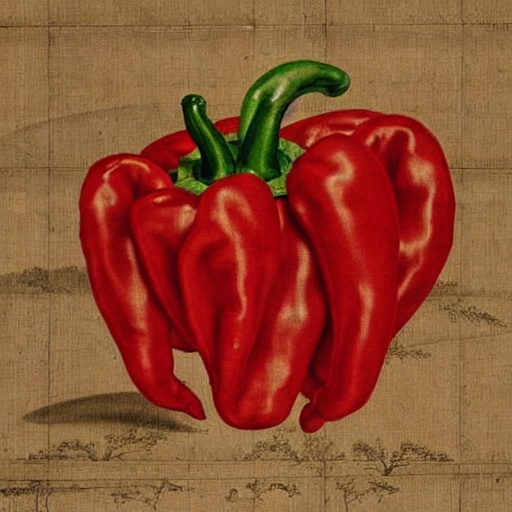} &
\includegraphics[width=0.13\linewidth]{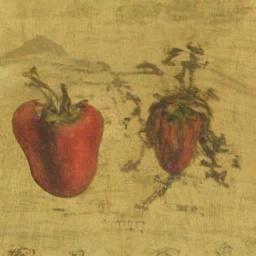} & \includegraphics[width=0.13\linewidth]{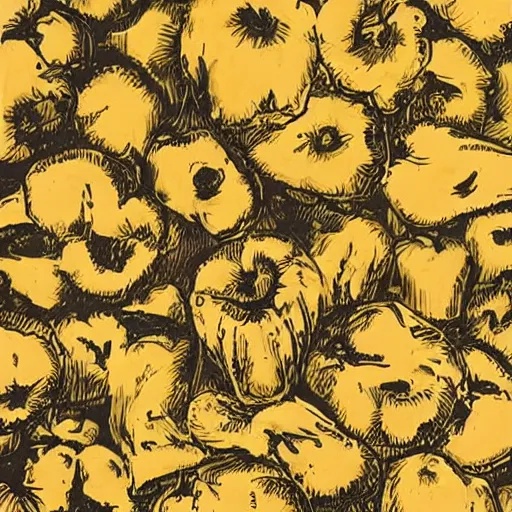} & \includegraphics[width=0.13\linewidth]{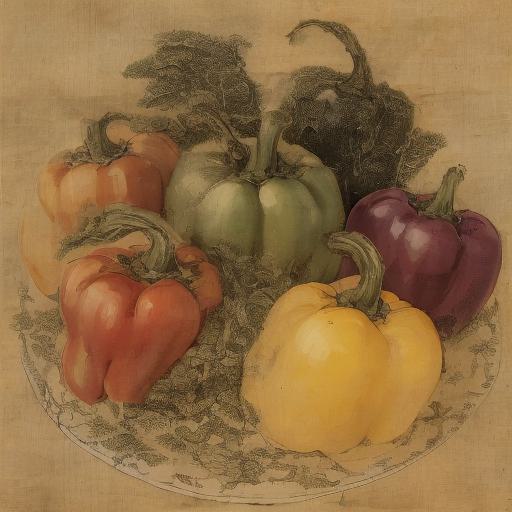}
\\

\includegraphics[width=0.13\linewidth]{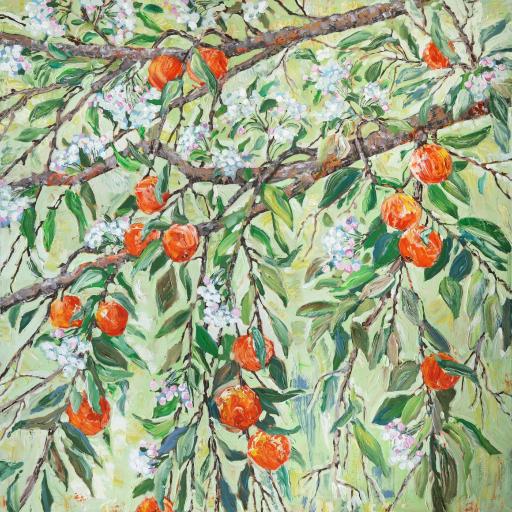} &
\includegraphics[width=0.13\linewidth]{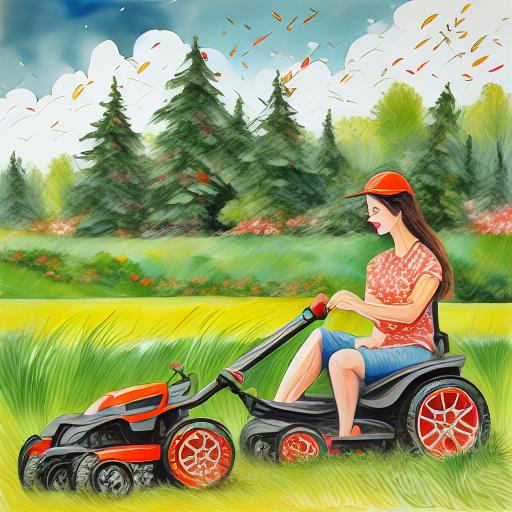} & 
\includegraphics[width=0.13\linewidth]{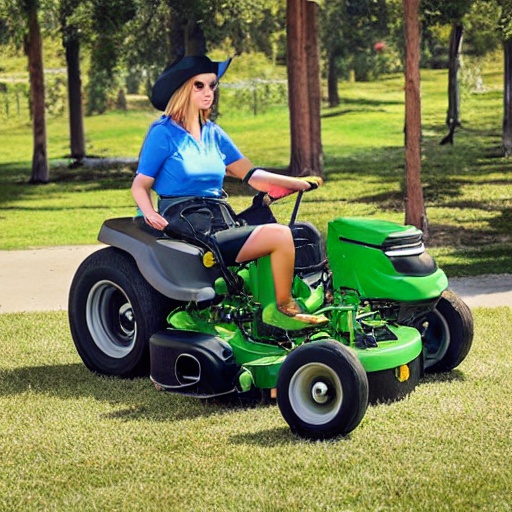} &
\includegraphics[width=0.13\linewidth]{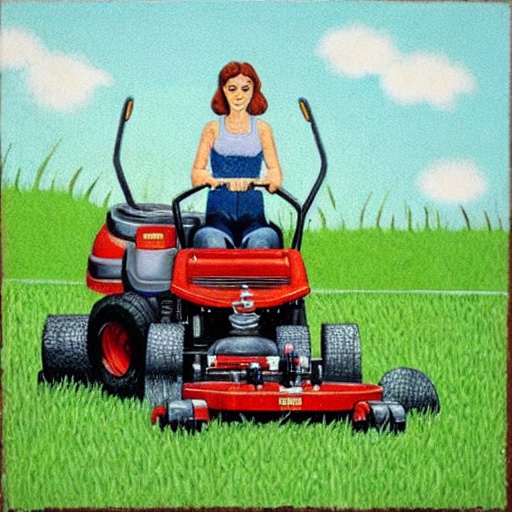} &
\includegraphics[width=0.13\linewidth]{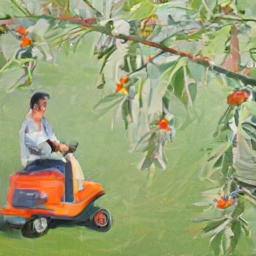} & \includegraphics[width=0.13\linewidth]{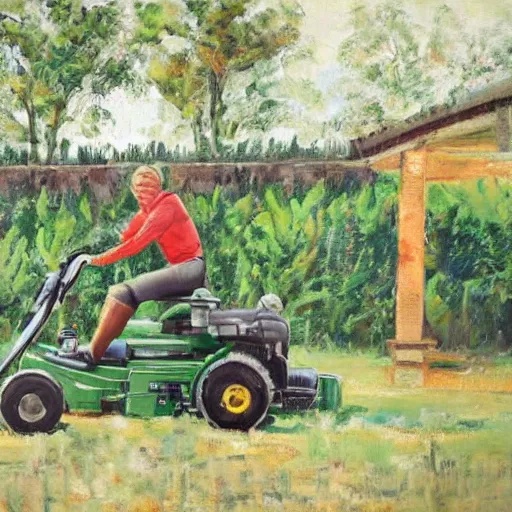} & \includegraphics[width=0.13\linewidth]{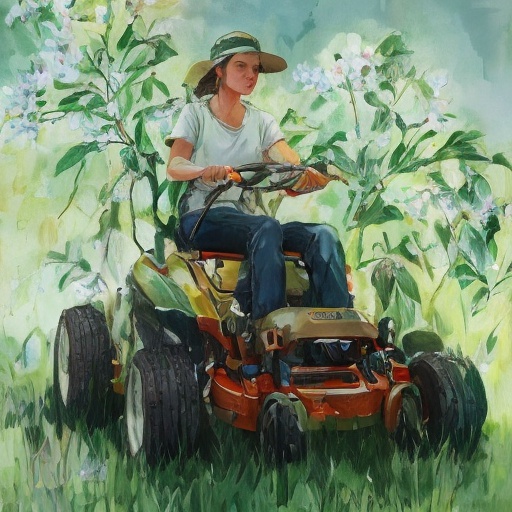}
\\

\includegraphics[width=0.13\linewidth]{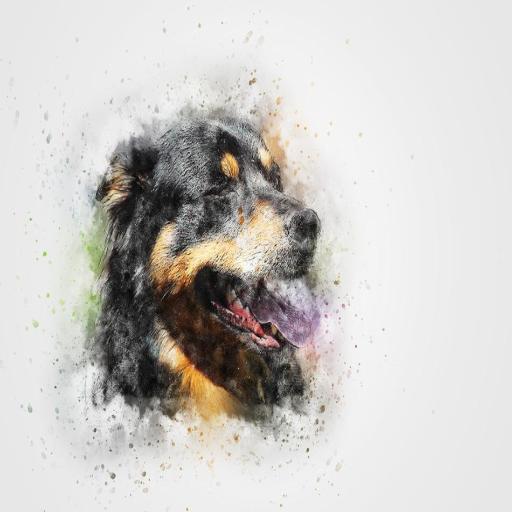} &
\includegraphics[width=0.13\linewidth]{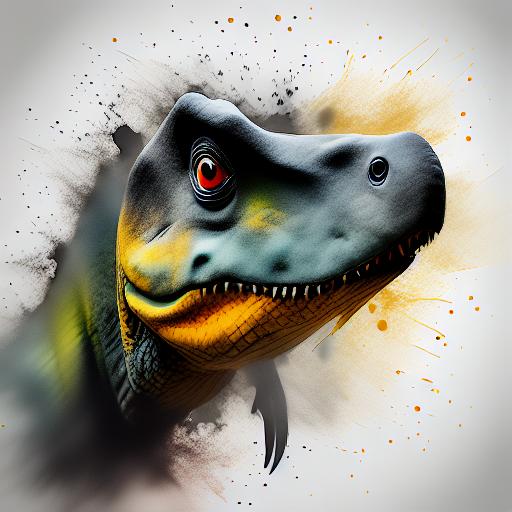} & 
\includegraphics[width=0.13\linewidth]{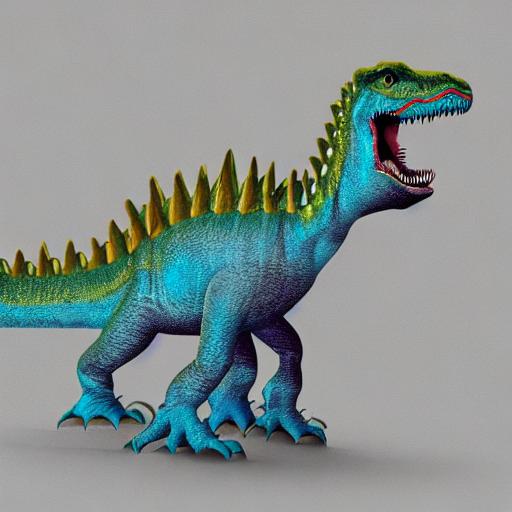} &
\includegraphics[width=0.13\linewidth]{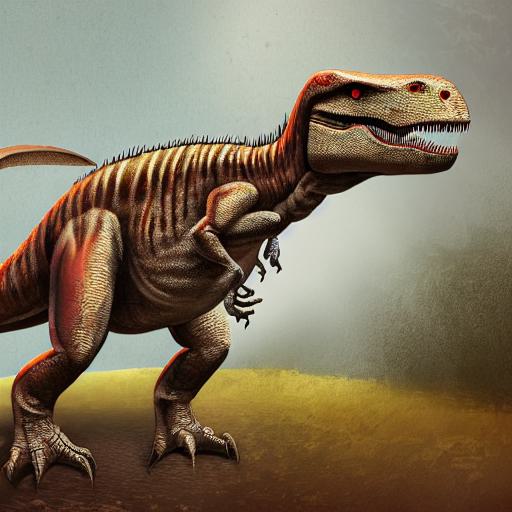} &
\includegraphics[width=0.13\linewidth]{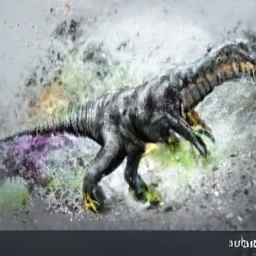} & \includegraphics[width=0.13\linewidth]{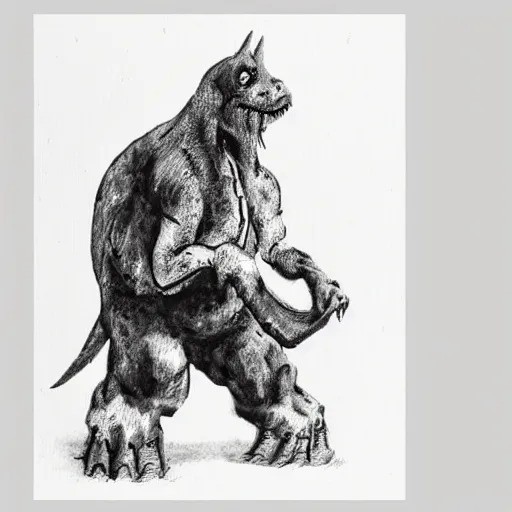} & \includegraphics[width=0.13\linewidth]{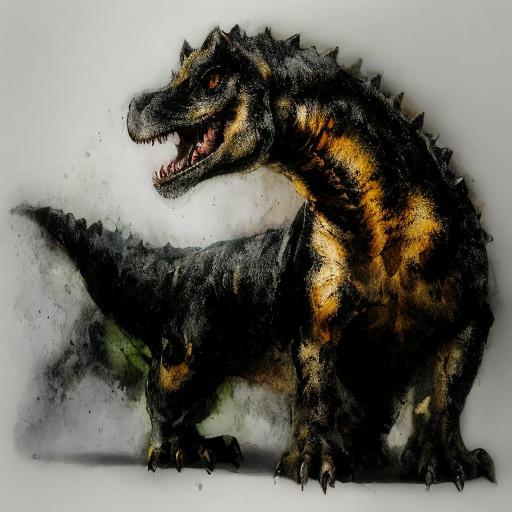}
\\

\includegraphics[width=0.13\linewidth]{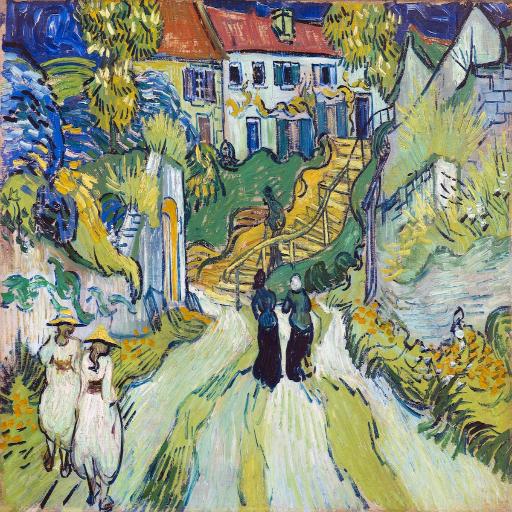} &
\includegraphics[width=0.13\linewidth]{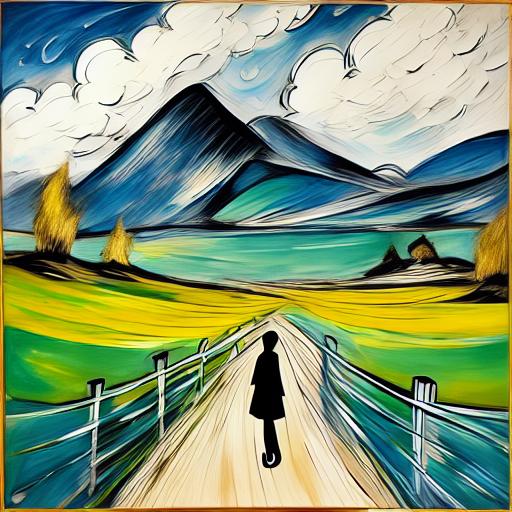} & 
\includegraphics[width=0.13\linewidth]{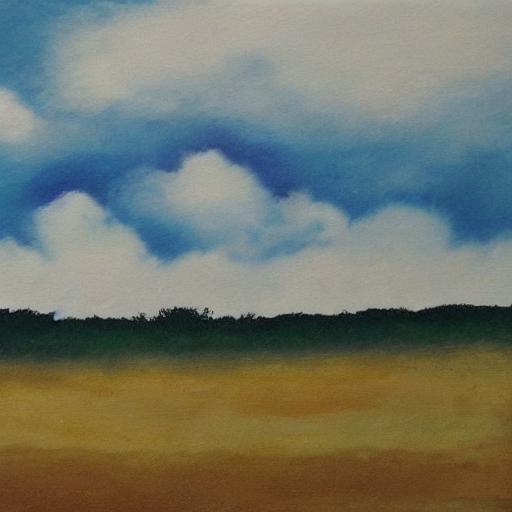} &
\includegraphics[width=0.13\linewidth]{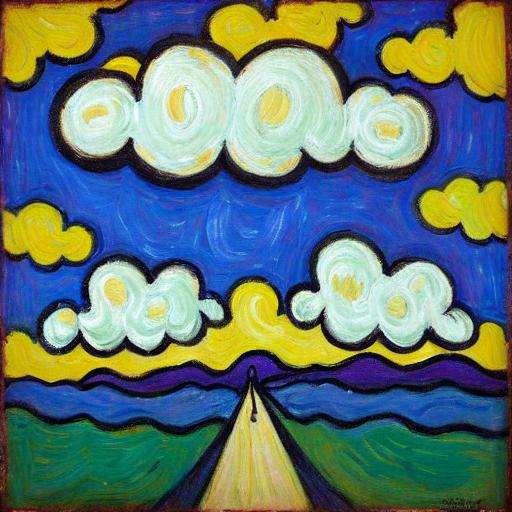} &
\includegraphics[width=0.13\linewidth]{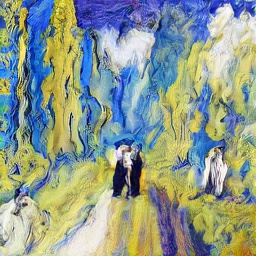} & \includegraphics[width=0.13\linewidth]{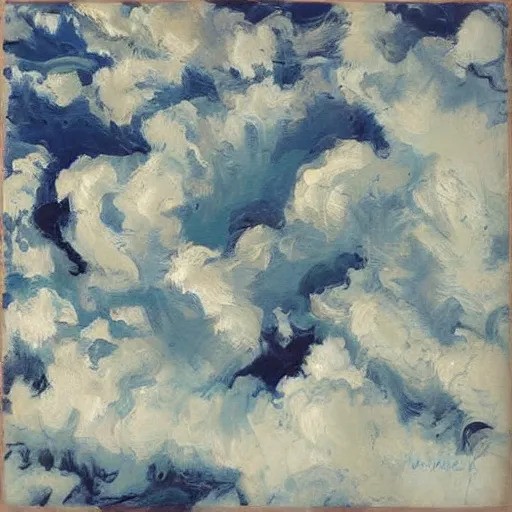} & \includegraphics[width=0.13\linewidth]{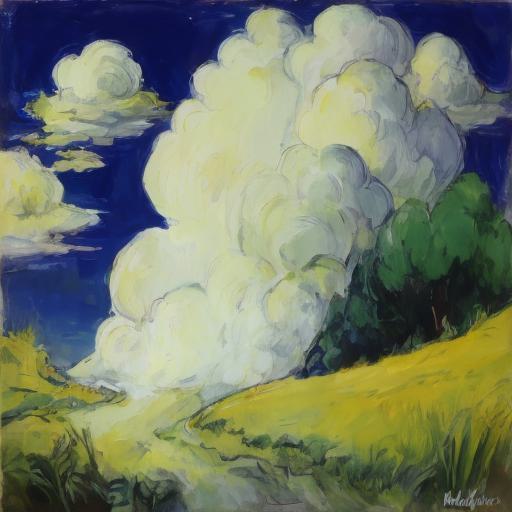}
\\

Style image & DEADiff & Custom-Diffusion & Dreamboot & StyleDrop & InstaStyle & Ours

\end{tabular}
\caption{\textbf{Qualitative comparison of stylized T2I generation on various style images.} The prompts for synthesis, listed from top to bottom, are: ``sweet peppers", ``woman driving lawn mower", ``dinosaur", and ``clouds". Our method effectively captures fine-grained style details, including color, textures, and so on.}
\label{fig_additional_comparison}
\end{figure*}

\begin{figure*}[!ht]
\centering
\begin{tabular}{c@{\hspace{0.3em}}c@{\hspace{0.3em}}c@{\hspace{0.3em}}c@{\hspace{0.3em}}c@{\hspace{0.3em}}c@{\hspace{0.3em}}c@{\hspace{0.3em}}}

\includegraphics[width=0.13\linewidth]{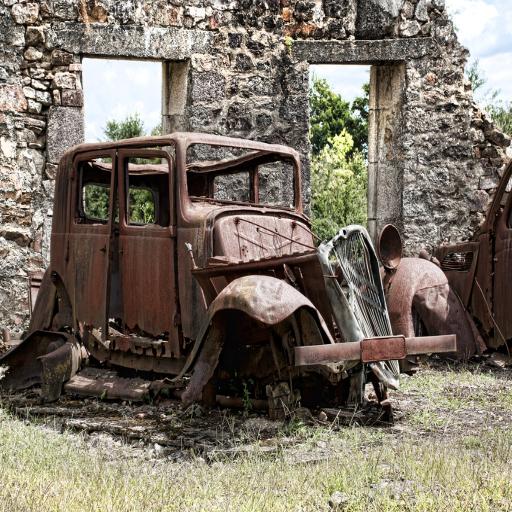} &
\begin{overpic}[width=0.13\linewidth]{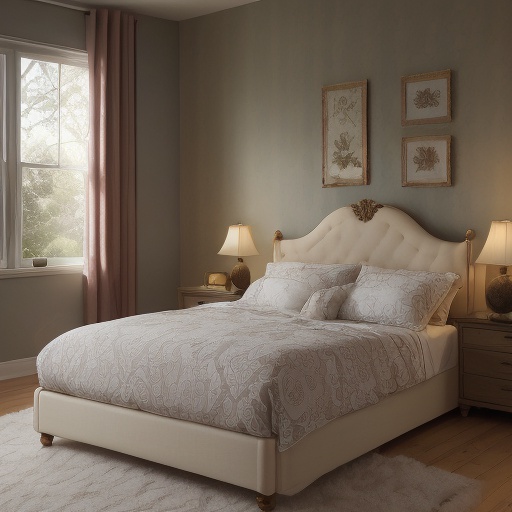}
  \put(10,10){\color{white}\bfseries\itshape\large bed}
\end{overpic} & 
\includegraphics[width=0.13\linewidth]{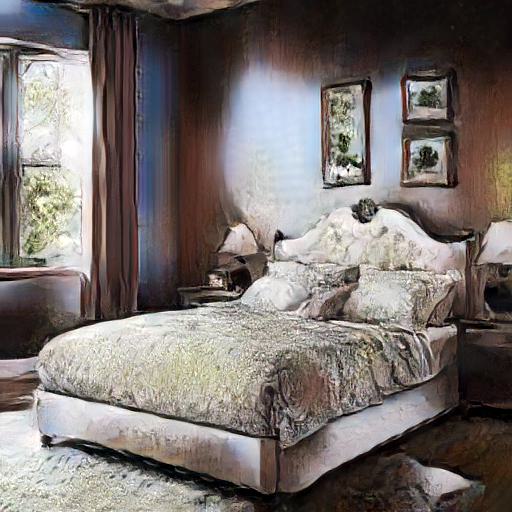} &
\includegraphics[width=0.13\linewidth]{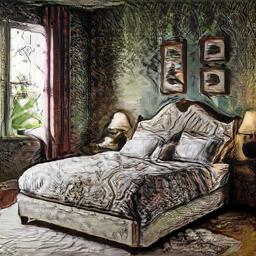} &
\includegraphics[width=0.13\linewidth]{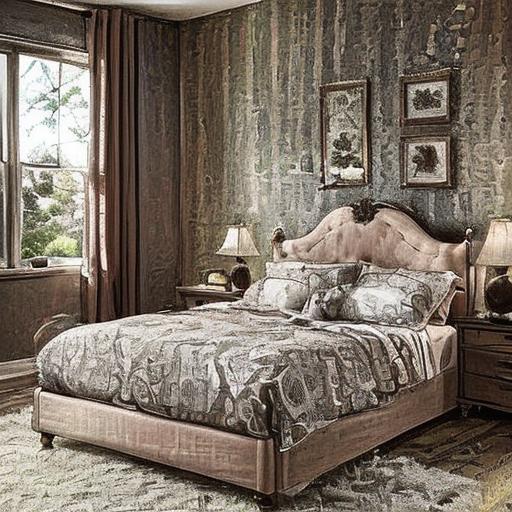} & \includegraphics[width=0.13\linewidth]{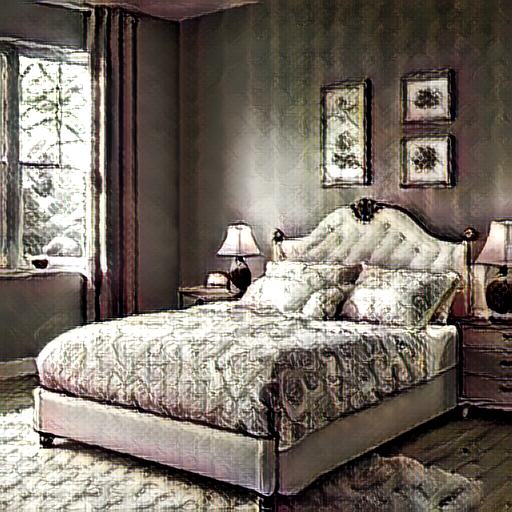} & \includegraphics[width=0.13\linewidth]{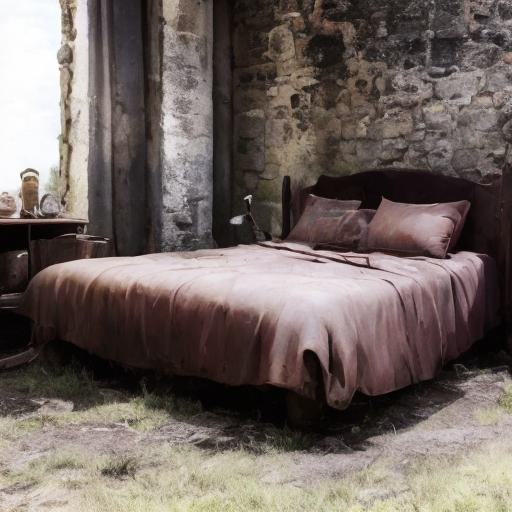}
\\

\includegraphics[width=0.13\linewidth]{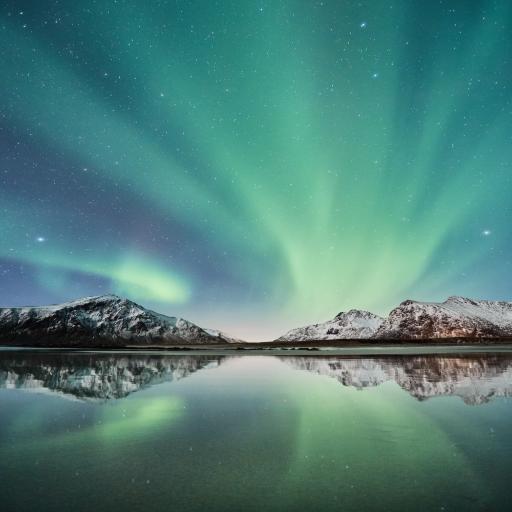} &
\begin{overpic}[width=0.13\linewidth]{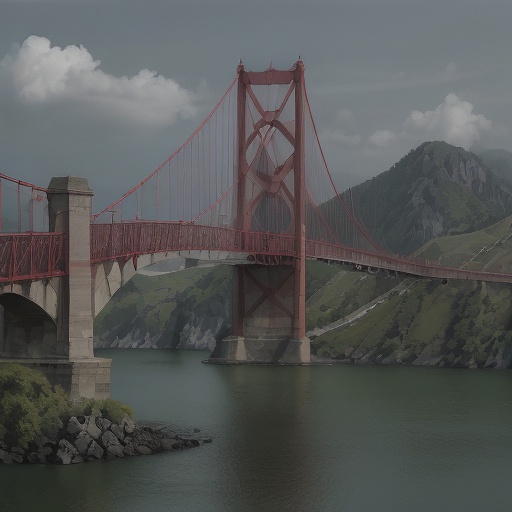}
  \put(10,10){\color{white}\bfseries\itshape\large bridge}
\end{overpic} & 
\includegraphics[width=0.13\linewidth]{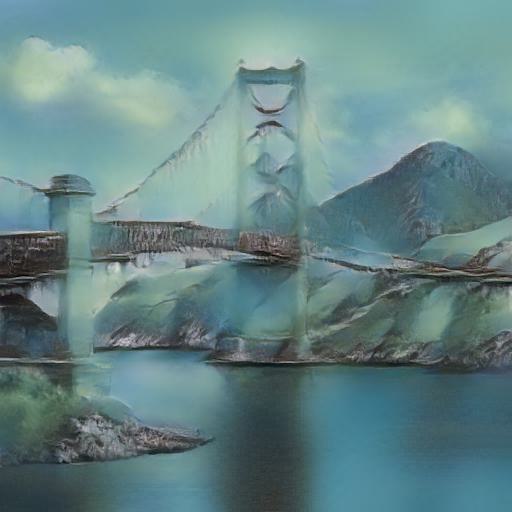} &
\includegraphics[width=0.13\linewidth]{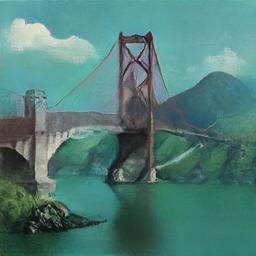} &
\includegraphics[width=0.13\linewidth]{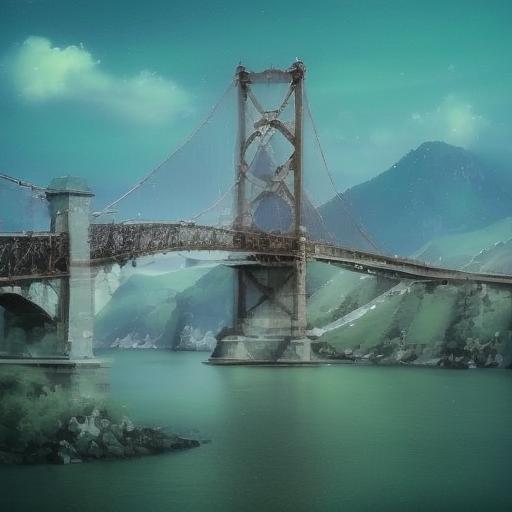} & \includegraphics[width=0.13\linewidth]{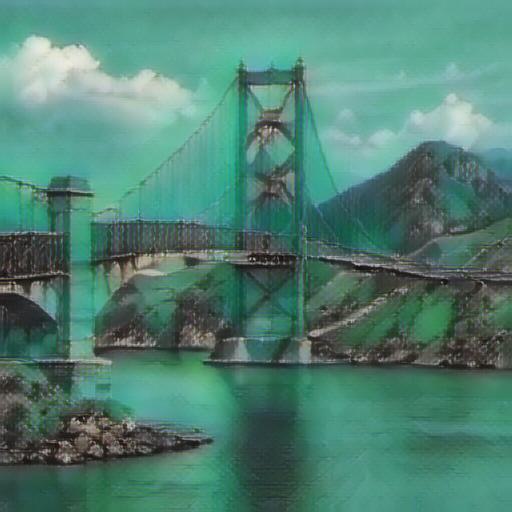} & \includegraphics[width=0.13\linewidth]{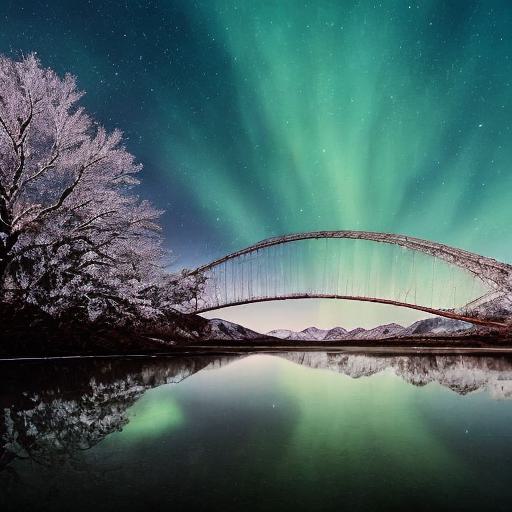}
\\

\includegraphics[width=0.13\linewidth]{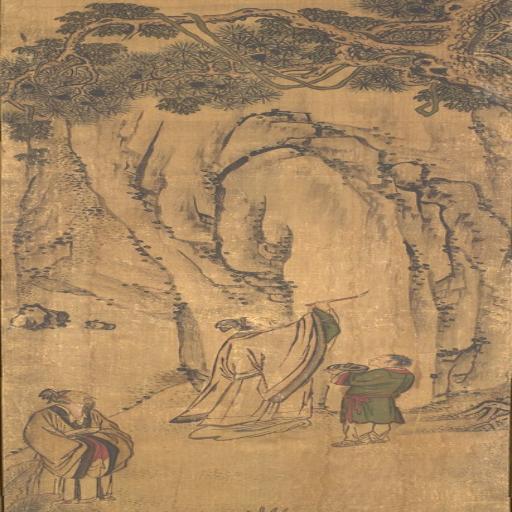} &
\begin{overpic}[width=0.13\linewidth]{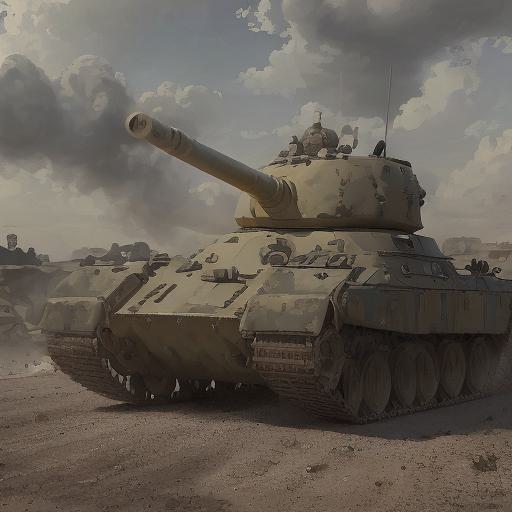}
  \put(10,10){\color{white}\bfseries\itshape\large tank}
\end{overpic} & 
\includegraphics[width=0.13\linewidth]{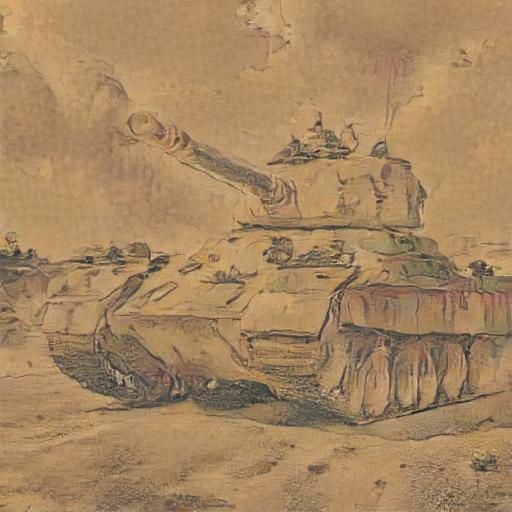} &
\includegraphics[width=0.13\linewidth]{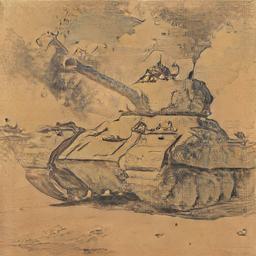} &
\includegraphics[width=0.13\linewidth]{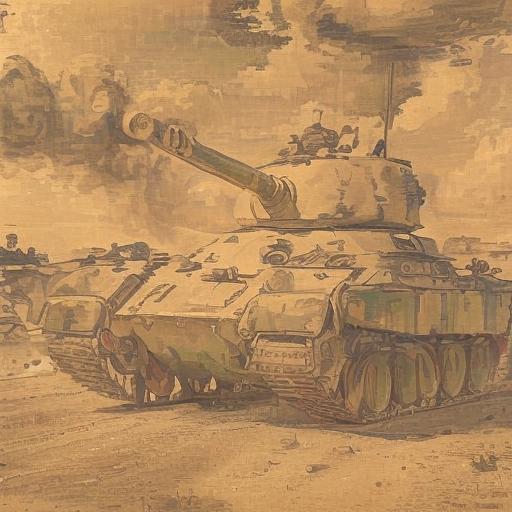} & \includegraphics[width=0.13\linewidth]{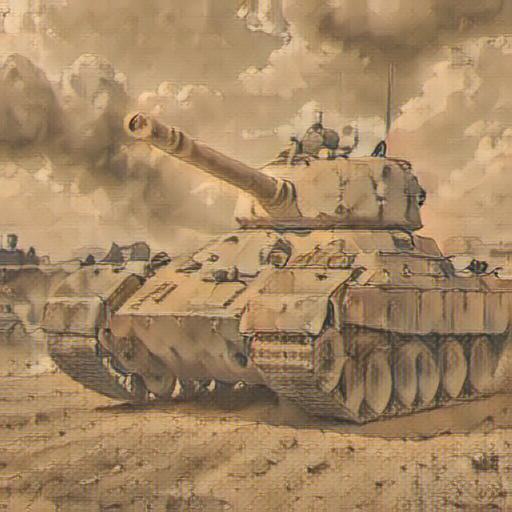} & \includegraphics[width=0.13\linewidth]{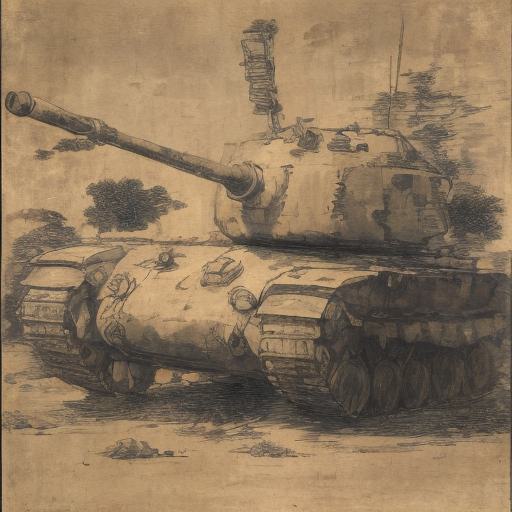}
\\

Style image & Content & AesPA-Net & CAST & StyleID & CCP & Ours

\end{tabular}
\caption{\textbf{Qualitative comparison with style transfer methods.} Content images, displayed in the second column, are used by style transfer methods, whereas our method relies solely on the related prompts shown in white. Despite using only textual prompts to represent content, our method achieves comparable performance in content fidelity while demonstrating superior capture of style patterns.}
\vspace{-0.9cm}
\label{fig_additiaonl_comparison_with_style_transfer}
\end{figure*}

We present additional visual comparison results with state-of-the-art methods. As shown in Fig.~\ref{fig_additional_comparison} and Fig.~\ref{fig_comparison}, Custom Diffusion often struggles to capture the style patterns of style images. DreamBooth and StyleDrop, perform slightly better in preserving style patterns. The primary reason for the limited performance of all four methods on stylized T2I is that, in our scenario, only a single style image is available for fine-tuning, making it difficult for these methods to effectively capture the unique style information. The improved performance of DEADiff can be attributed to its well-designed training dataset. Among the compared methods, InstaStyle achieves the best results in capturing style, mainly due to its use of DDIM Inversion to generate multiple style-consistent images for fine-tuning. However, its performance heavily depends on the accuracy of DDIM Inversion reconstruction and is prone to producing images with structural inaccuracies. In contrast, our method can generate stylized images with fine-grained style details and higher fidelity, all without any fine-tuning or additional optimization. Fig.~\ref{fig_comparison_with_style_transfer} and Fig.~\ref{fig_additiaonl_comparison_with_style_transfer} show a qualitative comparison with style transfer methods. Our method demonstrates comparable performance in terms of content fidelity and excels at preserving the style details of the reference image.



\end{document}